\documentclass{article}

\usepackage{microtype}
\usepackage{graphicx}
\usepackage{subfigure}
\usepackage{booktabs} %
\usepackage{xcolor}
\usepackage{bm}
\usepackage{multirow}
\usepackage{bigstrut, tabularx, multirow, makecell, diagbox}
\usepackage{nicefrac}

\usepackage{amsmath,amsfonts,amsthm}

\usepackage{hyperref}

\usepackage[accepted]{icml2021}
\usepackage{etoolbox}
\newcommand{\algorithmicdoinparallel}{\textbf{do in parallel}}
\makeatletter
\AtBeginEnvironment{algorithmic}{%
  \newcommand{\FORP}[2][default]{\ALC@it\algorithmicfor\ #2\ %
    \algorithmicdoinparallel\ALC@com{#1}\begin{ALC@for}}%
}
\makeatother

\usepackage{microtype}
\usepackage{lipsum}
\usepackage{graphicx}
\usepackage{subfigure}
\usepackage{booktabs} %
\usepackage{amssymb,xcolor,amsthm}
\usepackage{theoremref}
\usepackage{stmaryrd}
\usepackage{tabularx}
\usepackage[export]{adjustbox}

\newcommand{\mbf}[1]{\mathbf{#1}}

\newcommand*{\tran}{^{\mkern-1.5mu\mathsf{T}}}
\newcommand{\mbb}[1]{\mathbb{#1}}
\newcolumntype{C}[1]{>{\centering\arraybackslash}m{#1}}
\newcolumntype{P}[1]{>{\centering\arraybackslash}p{#1}}

\newcommand{\mcal}{\mathcal}
\newcommand{\norm}[1]{\left\lVert#1\right\rVert}

\newtheorem{proposition}{Proposition}

\newtheorem{innercustomprop}{Proposition}
\newenvironment{customprop}[1]
{\renewcommand\theinnercustomprop{#1}\innercustomprop}
{\endinnercustomprop}

\definecolor{skyblue}{HTML}{1e7aeb}

\newcommand{\be}{\begin{equation}}
\newcommand{\ee}{\end{equation}}
\definecolor{Gray}{gray}{0.85}
\definecolor{LightCyan}{rgb}{0.88,1,1}

\makeatletter
\usepackage{xspace} 
\def\@onedot{\ifx\@let@token.\else.\null\fi\xspace}
\DeclareRobustCommand\onedot{\futurelet\@let@token\@onedot}

\newcommand{\figref}[1]{Fig\onedot~\ref{#1}}
\newcommand{\algref}[1]{Algorithm~\ref{#1}}
\newcommand{\eqnref}[1]{Eq\onedot~\eqref{#1}}
\newcommand{\secref}[1]{Section~\ref{#1}}
\newcommand{\tabref}[1]{Tab\onedot~\ref{#1}}

\newcommand{\propref}[1]{Proposition~\ref{#1}}

\newcommand{\appref}[1]{Appendix~\ref{#1}}
\newcommand{\bfx}{\mathbf{x}}

\newcommand{\bfW}{\mathbf{W}}

\newcommand{\bfb}{\mathbf{b}}

\newcommand{\bfu}{\mathbf{u}}

\newcommand{\bfs}{\mathbf{s}}

\def\eg{\emph{e.g}\onedot}

\def\ie{\emph{i.e}\onedot}

\def\cf{\emph{cf}\onedot}

\def\wrt{w.r.t\onedot}

\icmltitlerunning{Accelerating Feedforward Computation via Parallel Nonlinear Equation Solving}

\begin{document}

\twocolumn[
\icmltitle{Accelerating Feedforward Computation via \\ Parallel Nonlinear Equation Solving}
\icmlsetsymbol{equal}{*}

\begin{icmlauthorlist}
\icmlauthor{Yang Song}{s}
\icmlauthor{Chenlin Meng}{s}
\icmlauthor{Renjie Liao}{t,u}
\icmlauthor{Stefano Ermon}{s}
\end{icmlauthorlist}

\icmlaffiliation{s}{Computer Science Department, Stanford University.}
\icmlaffiliation{t}{Department of Computer Science, University of Toronto.}
\icmlaffiliation{u}{Vector Institute}

\icmlcorrespondingauthor{Yang Song}{yangsong@cs.stanford.edu}
\icmlcorrespondingauthor{Stefano Ermon}{ermon@cs.stanford.edu}

\icmlkeywords{Machine Learning, feedforward computation, autoregressive models, DenseNet, ICML}

\vskip 0.3in
]

\printAffiliationsAndNotice{}  %

\begin{abstract}

Feedforward computation, such as evaluating a neural network or sampling from an autoregressive model, is ubiquitous in machine learning.
The sequential nature of feedforward computation, however, requires a strict order of execution and cannot be easily accelerated with parallel computing. To enable parallelization, we frame the task of feedforward computation as solving a system of nonlinear equations. We then propose to find the solution using a Jacobi or Gauss-Seidel fixed-point iteration method, as well as hybrid methods of both. Crucially, Jacobi updates operate independently on each equation and can be executed in parallel. Our method is guaranteed to give exactly the same values as the original feedforward computation with a reduced (or equal) number of parallelizable iterations, and hence reduced time given sufficient parallel computing power. Experimentally, we demonstrate the effectiveness of our approach in accelerating (i) backpropagation of RNNs, (ii) evaluation of DenseNets, and (iii) autoregressive sampling of MADE and PixelCNN++, with speedup factors between 2.1 and 26 under various settings.

\end{abstract}
\section{Introduction}

With sufficient parallel computing resources, we can certainly accelerate any algorithm with a parallelizable component. However, many machine learning algorithms heavily rely on a seemlingly non-parallelizable part---feedforward computation. %
To evaluate the output of a neural network, layers are computed one after the other in a feedforward fashion. 
To sample text from an autoregressive model, words are generated in sequence one by one. 
Because of the inherently sequential nature, it is difficult to directly perform feedforward computation in parallel---%
how can one output a label before any intermediate features are extracted, or generate the last word in a sentence before having seen the initial part?

At first sight, the idea of executing in parallel the various steps that comprise a feedforward computation procedure seems hopeless. Indeed, the task is clearly impossible in general. 
Machine learning workloads, however, have special properties that make the idea viable 
in some cases. 
First, \emph{computations are
numerical in nature}, and can tolerate small approximation errors. %
For example, it is acceptable if a faster method produces image samples at the cost of small imperceptible errors. 
Second, \emph{computations have been learned} from data rather than designed by hand.
As a result, they might involve unnecessary steps, and have dependencies between the various (sequential) stages that are weak enough to be ignored without significantly affecting the final results. Although we might not be able to explicitly characterize this structure of redundant dependencies, as long as it is present, we can design methods to take advantage of it.

Based on these insights, we propose an approach to accelerate feedforward computation with parallelism. Despite not beneficial for certain types of feedforward computation, it works well for many cases of practical interest in machine learning. %
Our key idea is to interpret feedforward computation as solving a triangular system of nonlinear equations, and use efficient numerical solvers to find the solution. This is advantageous because (i) many numerical equation solvers can be easily parallelized; and (ii) iterative numerical equation solvers generate a sequence of intermediate solutions of increasing quality, so we can use early stopping to trade off approximation error with computation time. In particular, we propose to find the solution of the triangular system 
using nonlinear Jacobi and Gauss-Seidel (GS) methods~\citep{ortega1970iterative}. Crucially, Jacobi iterations update each state independently and can be naturally executed in parallel. Moreover, we show feedforward computation corresponds to GS iterations, and can be combined with Jacobi iterations to build hybrid methods that interpolate between them.

We empirically demonstrate the effectiveness and flexibility of our proposed numerical equation solvers by showing accelerations for three representative applications: (i) the backpropagation procedure for training RNNs; (ii) the inference of neural networks like DenseNets~\citep{huang2017densely}; and (iii) ancestral sampling from autoregressive models such as MADE~\citep{germain2015made} and PixelCNN++~\citep{salimans2017pixelcnn++}. In particular, for the RNN model considered in our experiments, our new method reduces the training time by more than a factor of two. For DenseNet, our Jacobi-type methods lead to an estimated speedup factor of 2.1. For ancestral sampling from autoregressive models, we achieve 26 and 25 times speed up for MADE sampling on MNIST~\citep{lecun-mnisthandwrittendigit-2010} and CIFAR-10~\citep{krizhevsky2009learning} datasets; for PixelCNN++, we achieve 6.5 and 2.1 speedup factors respecitvely. Except for DenseNets where we simulate the performance due to computational constraints and implementation difficulties, all other results are measured with wall-clock time on a single GPU. This demonstrates that our methods not only perform well in the regime of massive parallel computing resources, but also have imminent practical values easily achievable with personal hardware.

\section{Background}\label{sec:background}
\subsection{Feedforward Computation}\label{sec:feedforward}
Consider the problem of computing, given an input $\bfu$, a sequence of \emph{states} $\bfs_1, \bfs_2, \cdots, \bfs_T$ 
defined by the following recurrence relation:
\begin{align}
    \bfs_t = h_t(\bfu, \bfs_{1:t-1}), & \quad 1 \leq t \leq T, \label{eqn:feedforward}
\end{align}
where $\{h_t\}_{t=1}^T$ are deterministic computable functions, and $\bfs_{1:t-1}$ is an abbreviation for $\bfs_1, \bfs_2, \cdots, \bfs_{t-1}$. %
From now on, we use $\bfs_{a:b}$ to denote $\bfs_a, \bfs_{a+1}, \cdots, \bfs_b$ where $a \le b$ and $a,b\in\mbb{N}^+$. %

Given implementations of the functions $\{h_t\}_{t=1}^T$, traditional \emph{feedforward computation} solves this problem by sequentially evaluating and memorizing $\bfs_t$, given $\bfu$ and the previously stored states $\bfs_{1:t-1}$.
Note that it cannot be na\"{i}vely parallelized across different time steps as each state $\bfs_t$ can only be obtained after we have already computed $\bfs_1, \cdots, \bfs_{t-1}$.

Feedforward computation is ubiquitous in machine learning. The following examples will appear in our experiments: (i) evaluating the output of a neural network layer by layer (neural network inference); (ii) back-propagating gradients from the loss function to weights (neural network training), and (iii) ancestral sampling from autoregressive models. For (i), $\bfu$ corresponds to the network input, and $\bfs_t$ denotes the activations of each layer; For (ii), $\bfu$ corresponds to the input and the activations stored during the forward pass, and $\bfs_t$ represents the gradient of the loss function \wrt each layer; For (iii), $\bfu$ is the latent state of a pseudo-random number generator, and $\bfs_t$ is the $t$-th dimension of the sample to be generated. See \appref{app:examples} for more detailed descriptions.

\subsection{Solving Systems of Nonlinear Equations}\label{sec:back_jacobi}

A system of nonlinear equations has the following form
\begin{equation}
f_i (x_1, x_2,\cdots, x_N) = 0, \quad i=1,2\cdots, N, \label{eqn:nonlinear}
\end{equation}
where $x_1, x_2, \cdots, x_N$ are unknown variables, and $f_1, f_2, \cdots, f_N$ are nonlinear functions. There are many effective numerical methods for solving systems of nonlinear equations. In this paper we mainly focus on 
nonlinear Jacobi and Gauss-Seidel methods, and refer 
to \cite{ortega1970iterative} for 
an excellent introduction to the field.

\subsubsection{Nonlinear Jacobi Iteration}\label{sec:back_jacobi}
To solve a system of 
equations like \eqnref{eqn:nonlinear}, iterative methods 
start from an initial guess $\bfx^0 \triangleq (x_1^0, x_2^0, \cdots, x_N^0)$ of the solution, and 
gradually improve 
it through fixed-point iterations. We let $\bfx^k = (x^k_1, x^k_2, \cdots, x^k_N)$ denote the solution obtained at the $k$-th iteration.
Given $\bfx^k$, the nonlinear Jacobi iteration produces $\bfx^{k+1}$ by solving each univariate equation for $i=1,2,\cdots, N$:
\begin{align}
    f_i(x_1^k, \cdots, x_{i-1}^k, x_i, x_{i+1}^k, \cdots, x_N^k) = 0\label{eqn:jacobi}
\end{align}
for $x_i$. We then set $x_i^{k+1} = x_i$ for all $i$. The process stops when it reaches a fixed point, or $\bfx^{k+1}$ is sufficiently similar to $\bfx^k$ as measured by the \emph{forward difference} $\norm{\bfx^{k+1} - \bfx^{k}} \leq \epsilon$, where $\epsilon > 0$ is a tolerance threshold. %
Crucially, all the $N$ univariate equations involved can be solved \emph{in parallel} since there is no dependency among them.

\subsubsection{Nonlinear Gauss-Seidel (GS) Iteration} 
Nonlinear Gauss-Seidel (GS) iteration is another iterative solver for systems of nonlinear equations. Similar to \eqnref{eqn:jacobi}, the $k$-th step of nonlinear GS is to solve
\begin{align}
    f_i(x_1^{k+1}, \cdots, x_{i-1}^{k+1}, x_i, x_{i+1}^k, \cdots, x_N^k) = 0 \label{eqn:gs}
\end{align}
for $x_i$ and to set $x_i^{k+1} = x_i$ for $i=1,2,\cdots,N$. The process stops when it reaches a fixed point, or $\norm{\bfx^{k+1} - \bfx^{k}} \leq \epsilon$. Different from \eqnref{eqn:jacobi}, GS updates leverage the new solutions as soon as they are available. This creates data dependency among adjacent univariate equations and therefore requires $N$ sequential computations to get $\bfx^{k+1}$ from $\bfx^k$. Assuming that each univariate equation of \eqnref{eqn:jacobi} and \eqnref{eqn:gs} takes the same time to solve, one GS iteration 
costs as much time as $N$ parallel Jacobi iterations.

Albeit one GS iteration involves sweeping over all variables and costs more compute than one Jacobi iteration, it can converge faster under certain cases, \eg, solving tridiagonal linear systems~\citep{young2014iterative}.

\section{Feedforward Computation as Equation Solving}
Our main insight is to frame a feedforward computation problem as solving a system of equations. This novel perspective enables us to use iterative solvers, such as nonlinear Jacobi and Gauss-Seidel methods, to parallelize and potentially accelerate traditional feedforward computation.

\subsection{Feedforward Computation Solves a Triangular System of Equations}
Given input $\bfu$, the recurrence relation among states $\bfs_1, \bfs_2, \cdots, \bfs_T$ in \eqnref{eqn:feedforward} can be explicitly expressed as the following system of nonlinear equations
\begin{equation}
h_t(\bfu, \bfs_{1:t-1}) - \bfs_{t} = 0, \quad t=1,2,\cdots, T \label{eqn:triangular}
\end{equation}
We can re-write \eqnref{eqn:triangular} as a systems of equations in the form of \eqnref{eqn:nonlinear} if we let $N=T$, $x_i \triangleq \bfs_i$, and $f_i(x_1,x_2, \cdots, x_T) \triangleq h_i(\bfu_i, \bfs_{1:i-1})-\bfs_i$, for $i=1,\cdots,N$. One unique property of these functions is that $f_i(\cdot)$ does not depend on $x_{i+1}, \cdots, x_N$, and therefore a recurrence relation corresponds to a \emph{triangular system} of nonlinear equations. Standard feedforward computation, as defined in \secref{sec:feedforward}, can be viewed as an iterative approach to solving the above triangular system of nonlinear equations.

\subsection{Jacobi Iteration for Recurrence Relations}
Any numerical equation solver can be employed to solve the system of nonlinear equations in \eqnref{eqn:triangular} and if converges, should return the same values as obtained by standard feedforward computation. As an example, we can use nonlinear Jacobi iterations to solve \eqnref{eqn:triangular}, as given in \algref{alg:jacobi}. Here we use $\bfs^{k}_{1:T}$ to denote the collection of all states at the $k$-th iteration, and choose $\epsilon > 0$ as a threshold for early stopping when $\norm{\bfs^k_{1:T} - \bfs^{k-1}_{1:T}} \leq \epsilon$, \ie, the \emph{forward difference} of states is small.

Although the nonlinear Jacobi iteration method is not guaranteed to converge to the correct solutions for general systems of equations~\citep{saad2003iterative}, it does converge for solving triangular systems. In particular, it is easy to conclude:
\begin{proposition}\label{prop:jacobi}
\algref{alg:jacobi} converges and yields the same result as standard feedforward computation in at most $T$ parallel iterations for any initialization of $\bfs_{1:T}^0$ if $\epsilon = 0$.
\end{proposition}
In the same vein, we can also apply nonlinear GS iterations to \eqnref{eqn:triangular}. Interestingly, running one iteration of GS is the same as performing standard feedforward computation and hence GS for triangular systems always converges in a single step, even though there is typically no convergence guarantee for more general systems of equations.

As already discussed in \secref{sec:background}, Jacobi iterations can exploit parallelism better than GS. Specifically, nonlinear Jacobi can complete $T$ iterations in parallel during which GS is only able to finish one iteration, if we assume that (i) the recurrence relation \eqnref{eqn:feedforward} can be evaluated using the same amount of time for all $t=1,\cdots,T$, and (ii) $T$ Jacobi updates can be done in parallel.
Thus, under these assumptions, \algref{alg:jacobi} can be much faster than the standard feedforward computation if the convergence of Jacobi iterations is fast. At least in the worst case, \algref{alg:jacobi} requires only $T$ iterations \emph{executed in parallel}, which takes the \emph{same wall-clock time} as one GS iteration (\ie, standard feedforward computation).

\subsection{Hybrid Iterative Solvers}
We can combine Jacobi and GS iterations to leverage advantages from both methods. The basic idea is to group states into blocks and view \eqnref{eqn:triangular} as a system of equations over these blocks. We can blend Jacobi and GS by first applying one of them to solve for the blocks, and then use the other to solve for individual states inside each block. Depending on which method is used first, we can define two different combinations dubbed Jacobi-GS and GS-Jacobi iterations respectively. 

\begin{algorithm}[!t]
   \caption{Nonlinear Jacobi Iteration}
   \label{alg:jacobi}
\begin{algorithmic}
   \STATE {\bfseries Input:} $\bfu$; $\epsilon$; $T$
   \STATE Initialize $\bfs_1^0, \bfs_2^0, \cdots, \bfs_T^0$ and set $k \gets 0$
   \REPEAT
   \STATE $k \gets k + 1$
   \FORP{$t=1$ {\bfseries to} $T$}
    \STATE $\bfs_t^k \gets h_t(\bfu, \bfs_{1:t-1}^{k-1})$
   \ENDFOR
   \UNTIL{$k = T$ \OR $\norm{\bfs^{k}_{1:T} - \bfs^{k-1}_{1:T}} \leq \epsilon$}
   \STATE {\bfseries return} $\bfs_1^k, \bfs_2^k, \cdots, \bfs_T^k$
\end{algorithmic}
\end{algorithm}
\begin{algorithm}[!ht]
   \caption{Nonlinear Jacobi-GS Iteration}
   \label{alg:jacobi_gs}
\begin{algorithmic}
   \STATE {\bfseries Input:} $\bfu$; $\mcal{B}_1, \mcal{B}_2, \cdots, \mcal{B}_M$; $\epsilon$; $T$
   \STATE Initialize $\bfs_1^0, \bfs_2^0, \cdots, \bfs_T^0$ and set $k \gets 0$
   \REPEAT
   \STATE $k \gets k + 1$
   \FORP{$i=1$ {\bfseries to} $M$}
        \STATE $\llbracket a, b \rrbracket \gets \mcal{B}_i$
        \FOR{$j \in \mcal{B}_i$}
            \STATE $\bfs_j^{k} \gets h_j(\bfu, \bfs_{1:a-1}^{k-1}, \bfs_{a:j-1}^{k})$
        \ENDFOR
   \ENDFOR
   \UNTIL{$k = M$ \OR $\norm{\bfs^{k}_{1:T} - \bfs^{k-1}_{1:T}} \leq \epsilon$}
   \STATE {\bfseries return} $\bfs_1^k, \bfs_2^k, \cdots, \bfs_T^k$
\end{algorithmic}
\end{algorithm}
\begin{algorithm}[th]
   \caption{Nonlinear GS-Jacobi Iteration}
   \label{alg:gs_jacobi}
\begin{algorithmic}
   \STATE {\bfseries Input:} $\bfu$; $\mcal{B}_1, \mcal{B}_2, \cdots, \mcal{B}_M$; $\epsilon$; $T$
   \STATE Initialize $\bfs_1, \bfs_2, \cdots, \bfs_T$
   \FOR{$i=1$ {\bfseries to} $M$}
        \STATE{Initialize $\bfs^{0}_j$ for all $j\in\mcal{B}_i$ and set $k\gets 0$}
        \STATE $\llbracket a, b \rrbracket \gets \mcal{B}_i$
        \REPEAT
            \STATE $k \gets k + 1$
            \FORP{$j \in \mcal{B}_i$}
                \STATE $\bfs_j^{k} \gets h_j(\bfu, \bfs_{1:a-1}, \bfs_{a:j-1}^{k-1})$
            \ENDFOR
        \UNTIL{$k = |\mcal{B}_i|$ \OR $\norm{\bfs_{\mcal{B}_i}^k - \bfs_{\mcal{B}_i}^{k-1}} \leq \epsilon$}
        \STATE $\bfs_{\mcal{B}_i} \gets \bfs_{\mcal{B}_i}^k$
   \ENDFOR
   \STATE {\bfseries return} $\bfs_1, \bfs_2, \cdots, \bfs_T$
\end{algorithmic}
\end{algorithm}

Suppose we use an integer interval $\mcal{B} = \llbracket a, b\rrbracket$ to represent a block of variables $\{ \bfs_a, \bfs_{a+1}, \cdots, \bfs_{b} \}$, and let $\{\mcal{B}_{1}, \mcal{B}_2, \cdots, \mcal{B}_M\}$ be a set of integer intervals that partitions $\llbracket 1, T\rrbracket$. We formally define Jacobi-GS in \algref{alg:jacobi_gs}, where $\bfs_{\mcal{B}}$ is a shorthand for $\{ \bfs_i \mid i \in \mcal{B} \}$. GS-Jacobi can be similarly defined and we provide its pseudo-code in \algref{alg:gs_jacobi}. Particularly, in Jacobi-GS (\algref{alg:jacobi_gs}), all $M$ blocks are updated in parallel and states within each block $\mcal{B}_i$ are updated sequentially based on the latest solutions. In GS-Jacobi (\algref{alg:gs_jacobi}), we sequentially update the $M$ blocks based on the latest solutions of previous blocks and the states within each block $\mcal{B}_i$ are updated in parallel.

Since \eqnref{eqn:triangular} is a triangular system of nonlinear equations, we have the following observation:
\begin{proposition}\label{prop:others}
For any initialization, Jacobi-GS (\algref{alg:jacobi_gs}) and GS-Jacobi (\algref{alg:gs_jacobi}) converge in at most $M$ block-wise iterations and yield the same results as obtained by standard feedforward computation if $\epsilon = 0$.
\end{proposition}
In summary, all the numerical equation solvers discussed above have guaranteed convergence in finite steps when solving our triangular systems of nonlinear equations in \eqnref{eqn:triangular}, and can thus act as valid alternatives to standard feedforward computation. Traditional asymptotic analysis of convergence rates is not applicable here, since the quotient convergence factor is undefined, and the root convergence factor is zero (per the definitions in \citet{ortega1970iterative}) when methods converge in finite steps.

\section{Accelerating~Feedforward~Computation}
\label{sec:which}

Below we discuss when Jacobi or hybrid methods can accelerate feedforward computation. We start with a computation model that is idealized but captures important practical aspects of Jacobi methods. The computation model assumes (i) for all $t=1,2,\cdots, T$, the recurrence relation \eqnref{eqn:feedforward} takes the same amount of time to compute for all values that $\bfs_{1:t-1}$ and $\bfu$ may take, and (ii) we have access to at least $T$ processors with the same computational power. For simplicity, we only count the computational cost of evaluating the recurrence relation given in \eqnref{eqn:feedforward} and ignore other potential costs that depend more on specific hardware implementation, such as data movements and synchronization.

We now analyze the advantages of various methods when the recurrence relations have different structures under the above computation model, and when the computation model is relaxed.

\subsection{When to Use the Jacobi Solver}
The above computation model has already been used several times to argue that $T$ parallel iterations of the Jacobi method costs the same wall-clock time as one sequential iteration of the GS method (\ie, the standard feedforward computation). According to \propref{prop:jacobi}, the Jacobi algorithm converges within $T$ parallel iterations. This implies that \emph{running \algref{alg:jacobi} is always faster or equally fast than standard feedforward computation (or GS)}. 

Since Jacobi iterations use more processors for parallel execution, it is necessary to understand when the speedup of Jacobi methods is worthwhile. To get some intuition, we first consider some typical examples where Jacobi iterations may or may not lead to compelling speedups with respect to Gauss-Seidel.

\textbf{Example 1: fully independent chains.~} The best case for Jacobi iteration is when for each $t=1,\cdots,T$, $\bfs_t = h_t(\bfu)$. For recurrent relations where different states are fully independent of each other, one parallel iteration of Jacobi suffices to yield the correct values for all states, whereas standard feedforward computation needs to compute each state sequentially. Parallelism in this case results in the maximum possible speedup factor of $T$.

\textbf{Example 2: chains with long skip connections.~} Here is a slightly worse, but still advantageous case for Jacobi iterations: each state only depends on far earlier states in the sequence via long skip connections. One simple instance is when $\bfs_1 = h_1(\bfu)$ and $\bfs_t = h_t(\bfu, \bfs_1)$ for $t > 1$. The Jacobi method needs only 2 parallel iterations to obtain the correct values of all intermediate states, which leads to a speedup factor of $T / 2$. We note that skip connections are commonly used in machine learning models, for example in ResNets~\citep{he2016deep}, DenseNets~\citep{huang2017densely}, and the computational graph of RNN backpropagation due to shared weights across time steps.

\textbf{Example 3: Markov chains.~} The worst case for Jacobi iterations happens when the recurrence relation is strictly Markov, \ie, $\bfs_1 = h_1(\bfu)$ and $\bfs_t = h_t(\bfs_{t-1})$ for $t > 1$. The Markov property ensures that when $t > 1$, the only way for $\bfs_t$ to be influenced by the input $\bfu$ is through computing $\bfs_{t-1}$. Therefore, as long as $\bfs_T$ depends on $\bfu$ in a non-trivial way, it will take at least $T$ parallel iterations for the Jacobi method to propagate information from $\bfu$ all the way to $\bfs_T$. In this case the running time of Jacobi matches that of GS or feedforward computation under our computation model.

In general, a recurrence relation can be represented as a directed acyclic graph (DAG) with $T+1$ nodes $\{\bfu, \bfs_1, \bfs_2, \cdots, \bfs_T\}$ to indicate computational dependency between states. The number of parallel iterations needed for the Jacobi method to converge is upper bounded by the \emph{critical path length}~\citep{kelley1959critical} (\ie, the length of the longest path between all pairs of nodes), whereas the number of iterations required for standard feedforward computation is always $T$. Therefore, \emph{Jacobi methods are better when the DAG has a smaller critical path length}. 

In the strict sense, DAGs of many feedforward processes in machine learning may not have a small critical path length. For example, DenseNets have a critical path length of $T$ since adjacent layers are connected, but empirically they enjoy substantial acceleration from Jacobi methods. This is because the influence of many connections is negligible (\eg, weights are small) and the DAG without these weak connections can have a much smaller effective critical path length. This frequently happens because models are learned rather than manually specified, and small numerical errors do not affect results.

We stress that all examples considered above are overly simplified for illustrative purposes. Our experiments in \secref{sec:exp} are on much more complicated tasks---neither RNN backpropagation, DenseNet evaluation, nor autoregressive sampling has a computational graph as simple as those examples. Empirically, we observe that Jacobi iterations have larger advantages when the computational graph of a machine learning task contains many long skip connections (\eg, DenseNets), but fall short when the computational graph is closer to a Markov chain (\eg, ResNets). Both are in agreement with the intuition given by our examples.

\subsection{When to Use Hybrid Solvers}
Our idealized computation model introduced at the beginning of this section assumes that we have $T$ parallel processors, and updates in the recurrence relation at $t=1,\cdots,T$ all have the same running time. When these assumptions do not hold, Jacobi-GS and GS-Jacobi are often more desirable than na\"{i}ve Jacobi iterations.%

First, when fewer than $T$ parallel processors are available, we cannot directly apply the Jacobi method. In contrast, both Jacobi-GS and GS-Jacobi require a smaller number of parallel processors equal to the number of blocks and the block size respectively, and can thus be tuned at will.

Second, when the computation time is non-uniform across different $t$, each parallel iteration of the Jacobi method will be bottlenecked by the slowest update across all time steps. One can use Jacobi-GS and GS-Jacobi to reduce this bottleneck, since the former can group different time steps so that each block takes roughly the same time to update, balancing the work load across different parallel processors; the latter can reduce the number of steps computed in parallel, leading to a smaller bottleneck during each GS update.

Third, when serial computation has unique advantages, the Jacobi method may have degraded performance as it is purely parallel. Under certain cases, the computation for $h_t(\bfu, \bfs_{1:t-1})$ can be cached to save the time for computing $h_{t+1}(\bfu, \bfs_{1:t})$ (\cf, \cite{ramachandran2017fast} for autoregressive models). This makes sequential computations faster than independent executions in parallel, and therefore reduces the cost-effectiveness of Jacobi methods compared to feedforward computation. In contrast, both Jacobi-GS and GS-Jacobi are more advantageous because the sequential GS iterations within and between blocks can also benefit from the faster serial computation brought by caches.

Finally, Jacobi-GS often converges faster than Jacobi even without the above considerations. For example, the ``block'' Jacobi method in the context of solving linear triangular systems is equivalent to our Jacobi-GS when applied to linear recurrence relations, and is shown to enjoy faster convergence than na\"{i}ve Jacobi iterations~\cite{chow2018using}.

\subsection{Practical Recommendations}
\textbf{Block size in hybrid solvers.~} When using hybrid methods, we should ensure that each block requires a comparable amount of computation. For Jacobi-GS, a larger block size requires fewer parallel computing units at the cost of slower running speed, while it is the opposite for GS-Jacobi. Users should balance this trade-off based on their goals and availability of computing units.

\textbf{Number of iterations.~} Determining the number of total iterations to run in advance is hard. Instead, we recommend an adaptive approach, where users stop the iteration once the forward difference (defined in \secref{sec:background}) is below a chosen tolerance value $\epsilon$ (see \algref{alg:jacobi}, \ref{alg:jacobi_gs} and \ref{alg:gs_jacobi}).

\section{Experiments}\label{sec:exp}

\begin{figure*}[!ht]
    \centering
    \subfigure[RNN training\label{fig:rnn}]{\includegraphics[width=0.34\linewidth, valign=c]{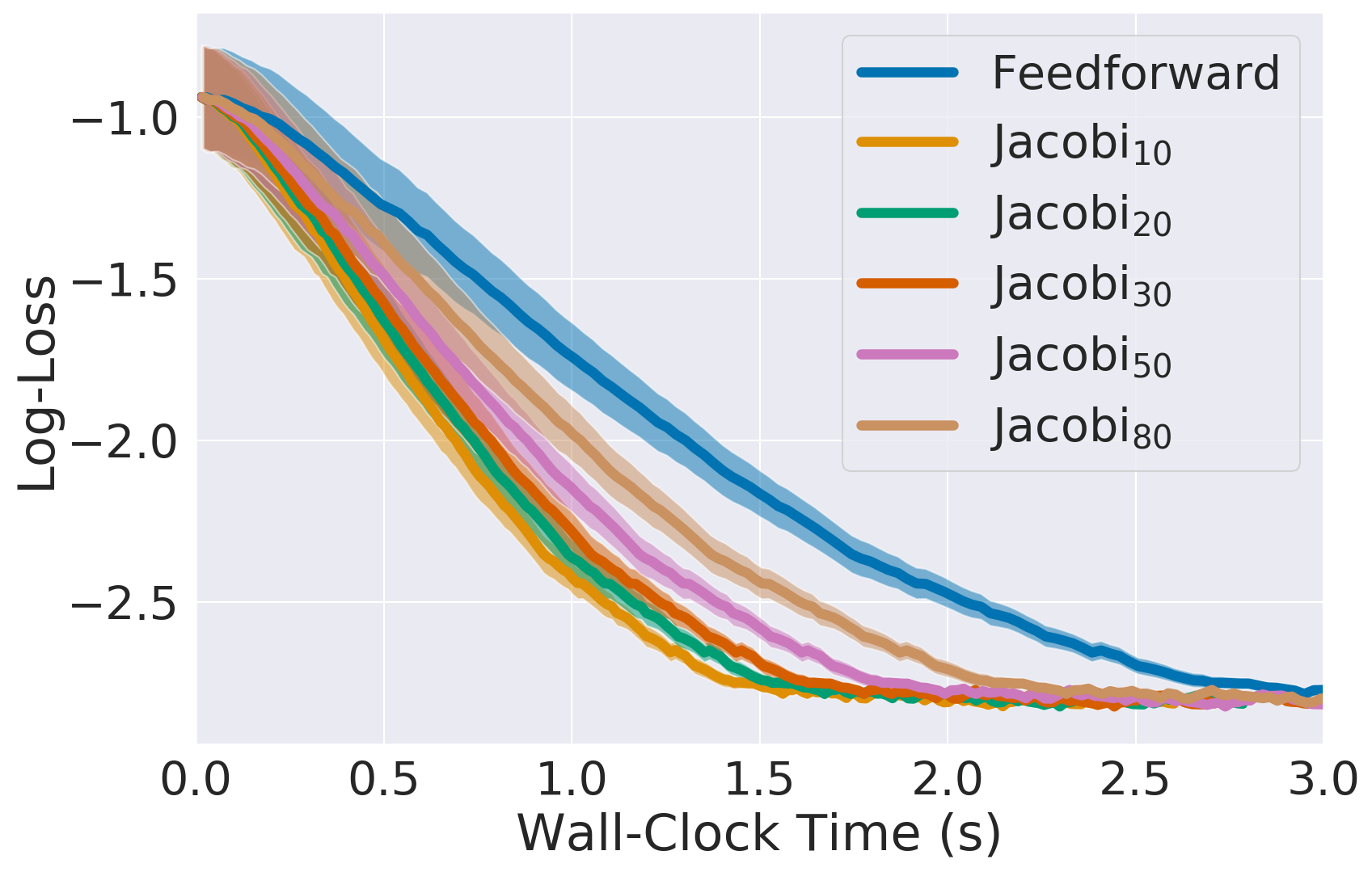}}%
    \subfigure[MADE sampling on MNIST\label{fig:made_mnist}]{\includegraphics[width=0.325\linewidth, valign=c]{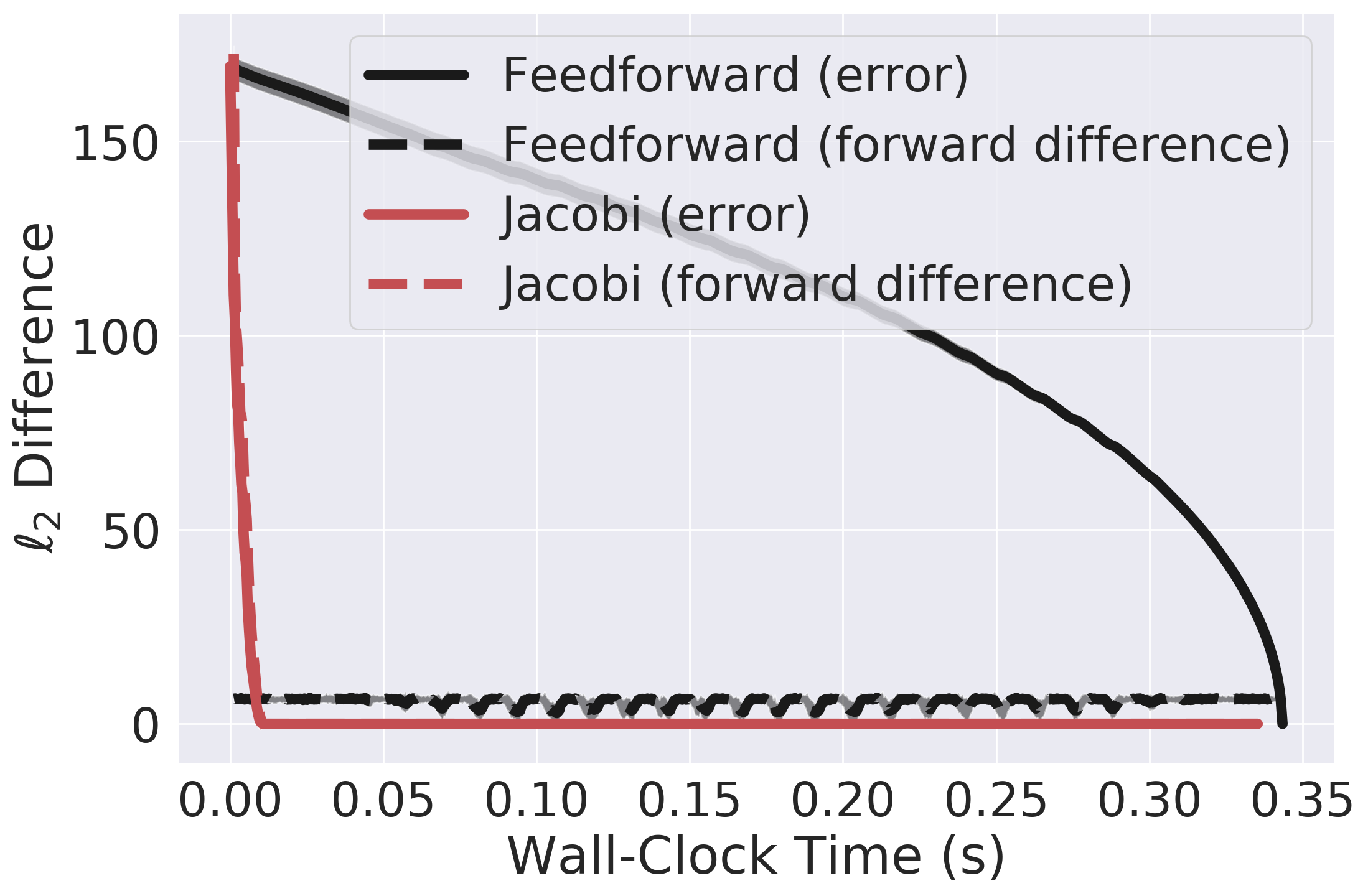}}
    \subfigure[PixelCNN++ sampling on MNIST\label{fig:pixelcnn_mnist}]{\includegraphics[width=0.325\linewidth, valign=c]{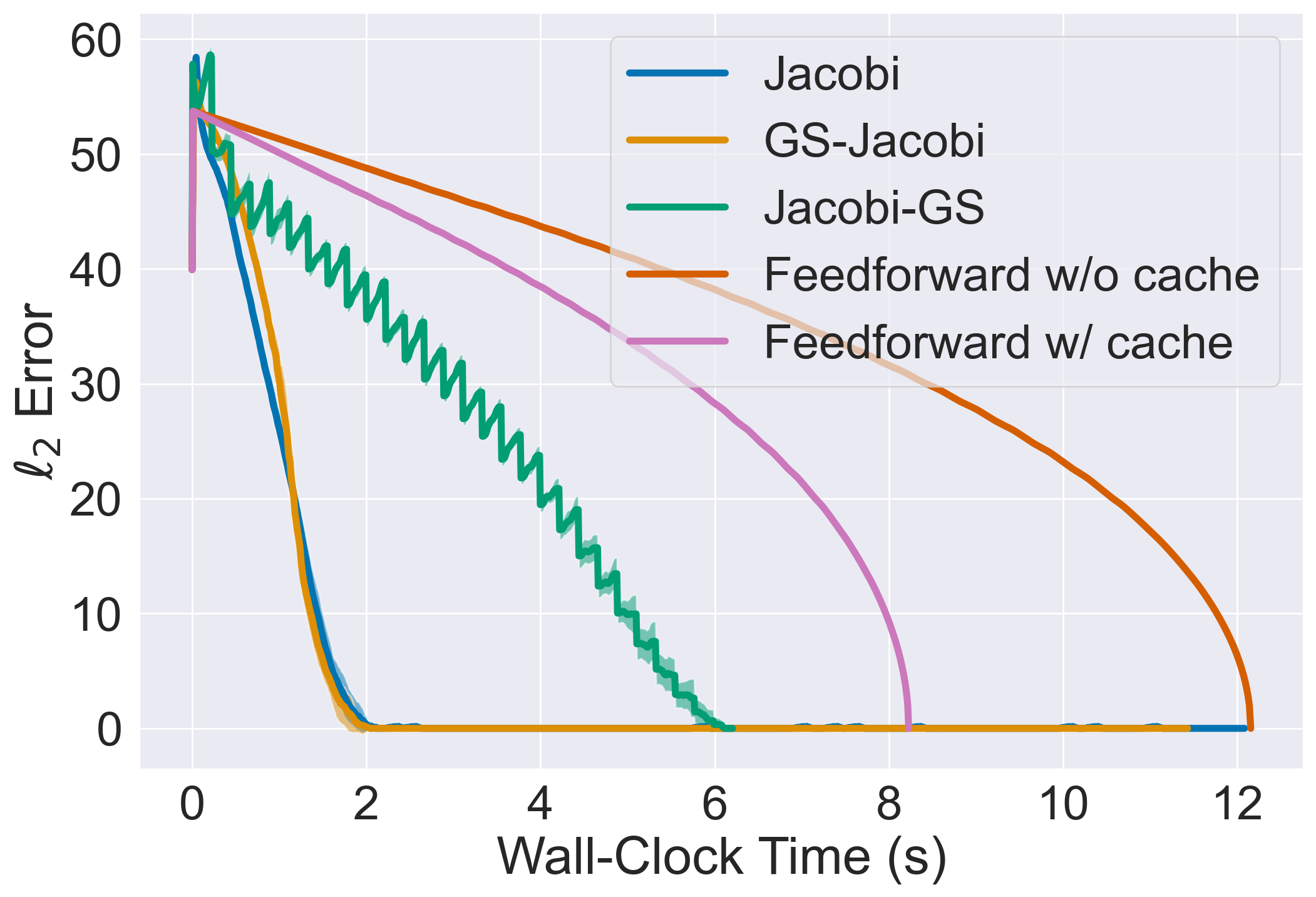}}\\
    \subfigure[DenseNet evalution\label{fig:densenet}]{\includegraphics[width=0.325\linewidth, valign=c]{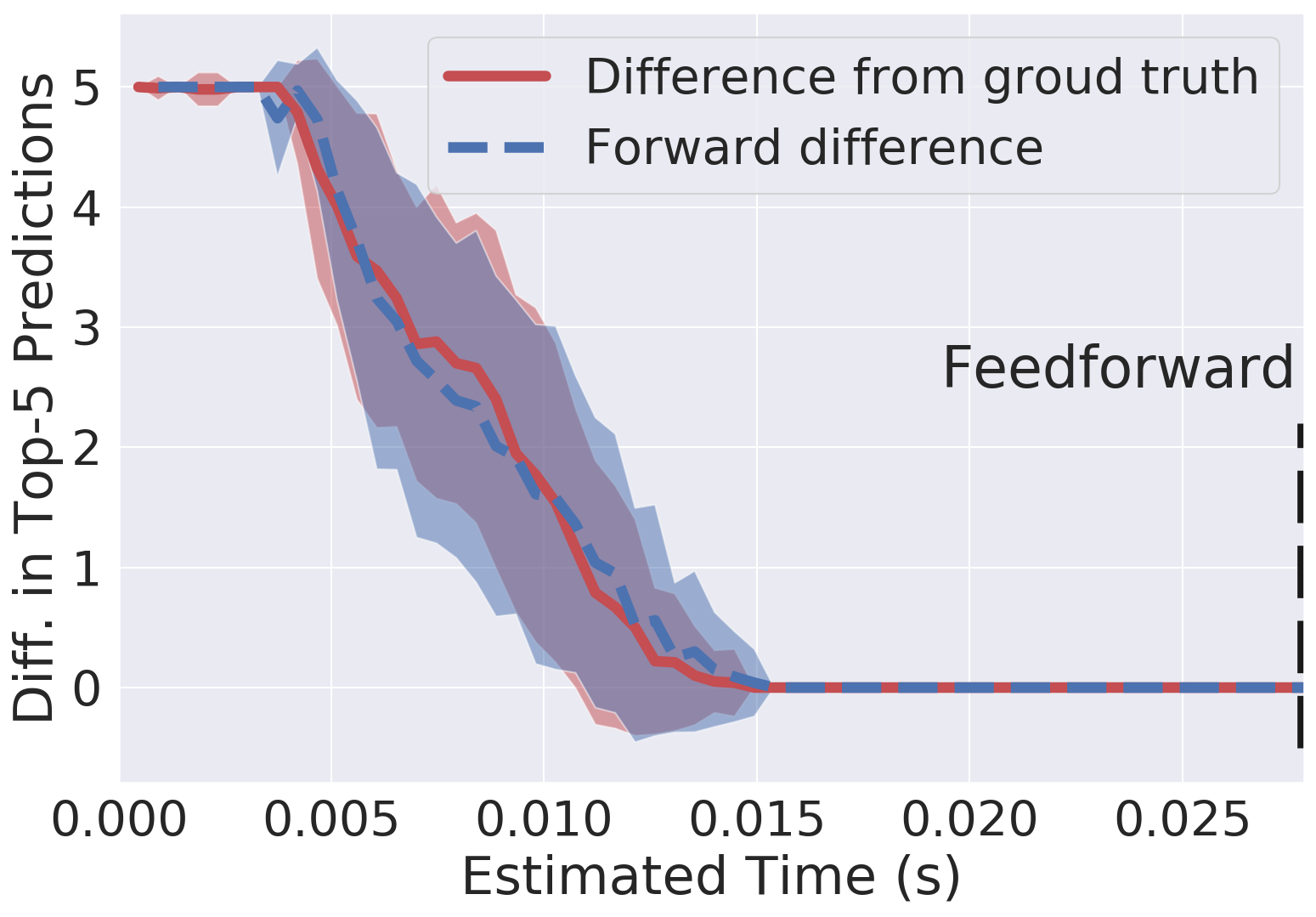}}
    \subfigure[MADE sampling on CIFAR-10\label{fig:made_cifar}]{\includegraphics[width=0.338\linewidth, valign=c]{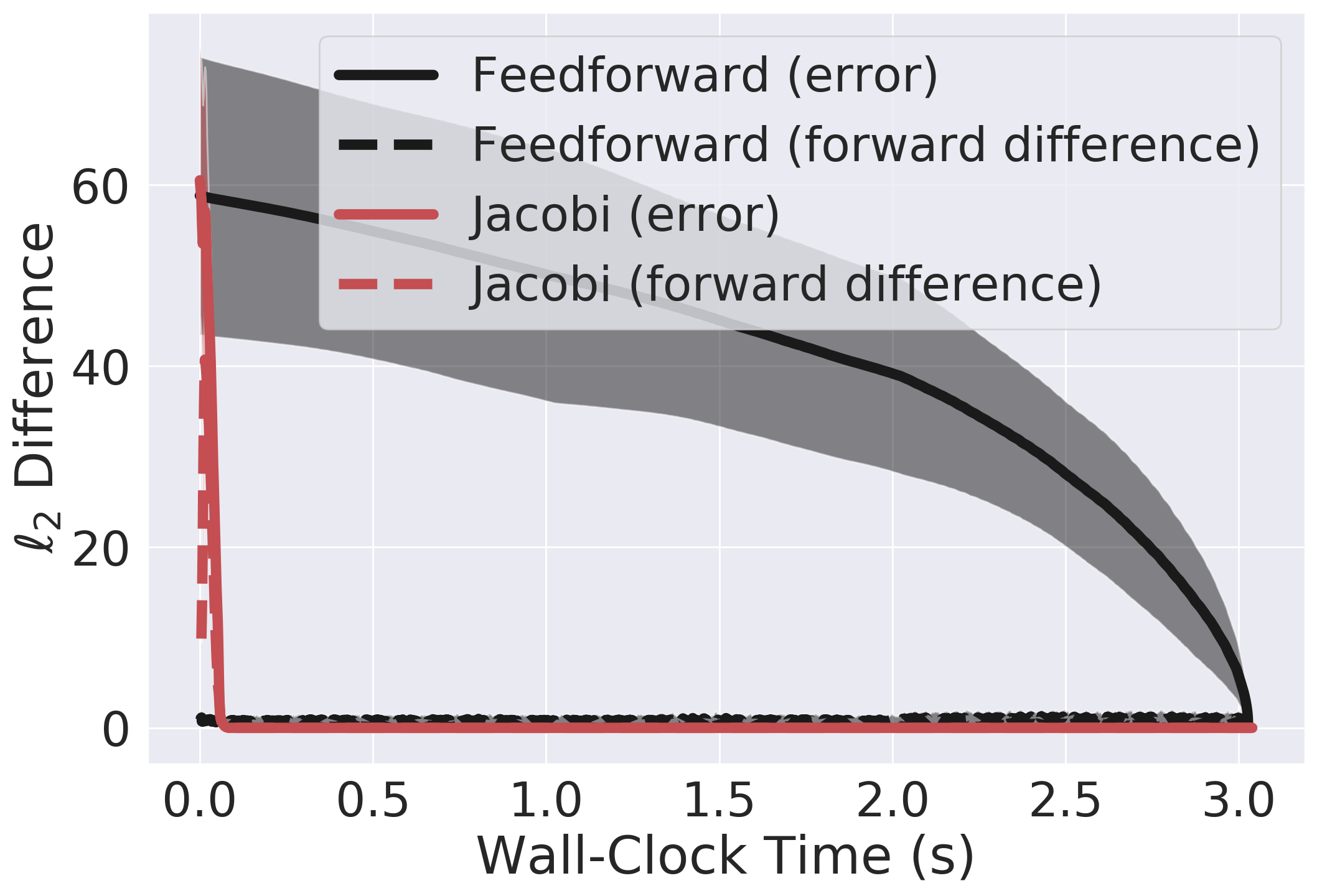}}%
    \subfigure[PixelCNN++ sampling on CIFAR-10\label{fig:pixelcnn_cifar}]{\includegraphics[width=0.33\linewidth, valign=c]{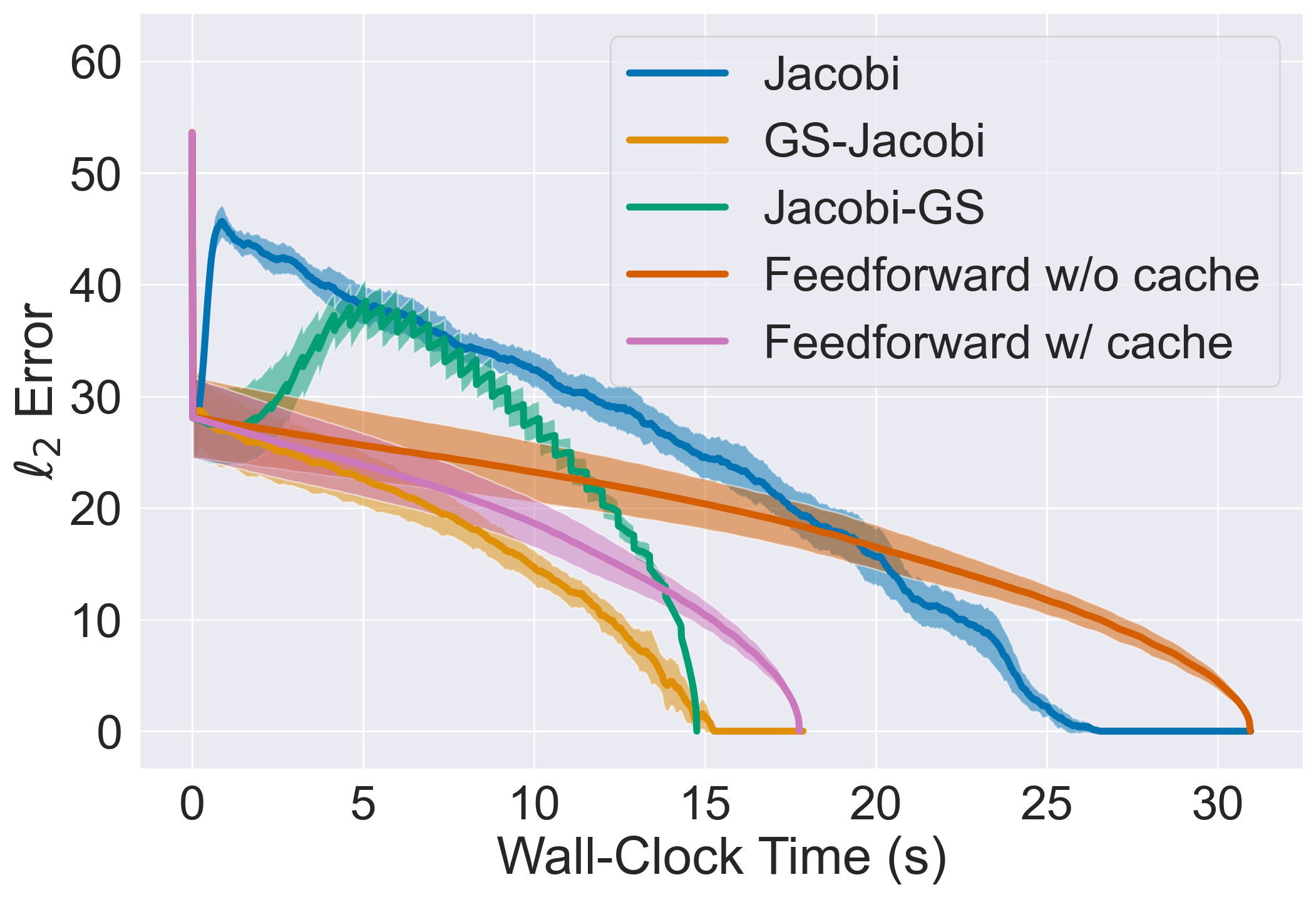}}

    \caption{(a) The performance of Jacobi iterations on accelerating RNN training. Here we use ``Jacobi$_n$'' to denote the Jacobi method truncated at the $n$-th iteration, and ``feedforward'' for standard backpropagation. All values are averaged over 10 runs and shaded areas denote $\nicefrac{1}{10}$ standard deviations. (d) The performance of Jacobi-GS on evaluating DenseNets. The y-axis represents the number of incorrect labels in top-5 predictions. The shaded areas represent standard deviations across 100 random input images. (b)(e) The performance of feedforward sampling vs. Jacobi iterations for MADE. The shaded areas represent standard deviations computed over 100 runs. (c)(f) Comparing different sampling algorithms for PixelCNN++. Results are averaged over 10 runs and shaded areas show standard deviations.}\label{fig:results}
\end{figure*}

Here we empirically verify the effectiveness of our proposed algorithms on 
(i) the backpropagation of RNNs, (ii) the evaluation of neural networks, and (iii) the ancestral sampling of deep autoregressive models. We report the speedups of our algorithms measured with wall-clock time on real hardware, except for the DenseNet experiment where we simulate the performance due to the difficulty of implementing our methods in current deep learning frameworks like PyTorch~\citep{paszke2019pytorch} and TensorFlow~\citep{tensorflow2015-whitepaper}. We provide the main experimental results with key details in this section, and relegate other details/results to \appref{app:exp_details}/\ref{app:exp_results}.

\subsection{Backpropagation of RNNs}
We consider accelerating the training procedure of a recurrent neural network (RNN) with Jacobi iterations. The backward pass of RNNs can  benefit from Jacobi-type approaches, because the loss function is connected to all time steps in the computation graph, and therefore gradient information can quickly flow from the final loss value to all hidden states with one Jacobi update.

To demonstrate this, we train a simple RNN with one hidden layer to fit sequences. The dataset is synthesized by flattening resized MNIST digits (resolution $10\times 10$). We report how the training loss decreases with respect to wall-clock time in \figref{fig:rnn}. Since the length of input sequences is fixed to $100$, there are a total of $100$ steps in the backward pass. We use ``Jacobi\textsubscript{$n$}'' to denote the Jacobi approach truncated at the $n$-th iteration ($n \leq 100$), and ``feedforward'' corresponds to the standard backpropagation algorithm. In \figref{fig:rnn_convergence} (see \appref{app:exp_results}), we show how Jacobi\textsubscript{$n$} converges to the true gradients with respect to $n$. We can trade-off between the accuracy and speed of gradient computation by tuning $n$. As demonstrated in \figref{fig:rnn}, Jacobi methods can reduce the training time by around a factor of two with a proper $n$.

\subsection{Evaluating DenseNets}
DenseNets~\citep{huang2017densely} are convolutional neural networks with a basic building block called the dense layer. Each dense layer contains two convolutions, and is connected to every other dense layer in a feedforward fashion. DenseNets are particularly suitable for Jacobi-type iterative approaches because information can quickly flow from input to output in one update via skip connections.

\paragraph{Setup.} We use a DenseNet-201 model pre-trained on ImageNet~\citep{ILSVRC15}. \emph{We define a state in the corresponding recurrence relation to be the feature maps of a convolutional layer}. We apply the Jacobi-GS method (\algref{alg:jacobi_gs}) to compute all states, where \emph{each dense layer (consisting of two states) is grouped as one block}. 
We empirically verify that evaluating each dense layer separately takes comparable running time on GPUs. Therefore, by arranging these dense layers as blocks, Jacobi-GS can have roughly balanced workload for parallel execution.

\paragraph{Performance Metrics.} For this task, full implementation of our algorithms will involve heterogeneous parallel execution of convolutional layers, which is not well supported by existing deep learning frameworks such as JAX~\citep{jax2018github}, PyTorch~\citep{paszke2019pytorch} or TensorFlow~\citep{tensorflow2015-whitepaper}. Therefore, we estimate the speedup for a real parallel implementation by simulating the performance of Jacobi-GS with a purely sequential implementation, assuming no overheads due to parallelism.
Specifically, we run each dense layer 10 times on the GPU and take the average to measure its wall-clock time, which we denote as $t_1, t_2, \cdots, t_{98}$, since there are 98 blocks in total. We then estimate one parallel iteration of Jacobi-GS with $\operatorname{max}_{1\leq i\leq98} t_i$, and the time for full feedforward computation with $\sum_{i=1}^{98} t_i$.

\paragraph{Results.}
We summarize the performance of Jacobi-GS in \figref{fig:densenet}. %
We plot the curves of both error and forward difference (defined in \secref{sec:back_jacobi}), measured using the number of different labels in top-5 predictions. The results indicate that forward differences closely trace the ground-truth errors and therefore can be reliably used as a stopping criterion. %
As shown in \figref{fig:densenet}, the estimated time for Jacobi-GS to converge is around \textbf{0.0131s}, which is \textbf{2.1} times faster than \textbf{0.0279s}, the estimated time needed for feedforward computation. Note that this is a theoretical speedup. The actual speedup might be smaller due to overheads of parallel execution. 

\if0
\begin{figure}
    \centering
    \includegraphics[width=0.6\linewidth]{figures/densenet4.png}
    \caption{The performance of Jacobi-GS on evaluating DenseNets. The y-axis represents the number of incorrect labels in top-5 predictions. The shaded areas represent standard deviations across 100 random input images. %
    }
    \label{fig:densenet}
\end{figure}
\fi

\subsection{Autoregressive Sampling}
We consider two popular autoregressive models for image generation: MADE~\citep{germain2015made} and PixelCNN++~\citep{salimans2017pixelcnn++}. Both generate images pixel-by-pixel in raster scan order, and thus every pixel forms a state in the corresponding recurrence relation of feedforward computation. 

\subsubsection{MADE}
For autoregressive sampling from MADE, each iteration of feedforward computation requires a forward propagation of the whole network, which equals the cost of one parallel Jacobi iteration. This means that sampling from MADE is a perfect use case for Jacobi iterations, where no extra parallelism is needed compared to na\"{i}ve feedforward computation.

\paragraph{Setup.} We compared Jacobi iteration against feedforward sampling for models trained on MNIST~\citep{lecun-mnisthandwrittendigit-2010} and CIFAR-10~\citep{krizhevsky2009learning} respectively. The experiments were repeated 100 times and we report the means and standard deviations measured in \textit{actual wall-clock time} on a single NVIDIA Titan Xp GPU, accounting for all the overheads. %

\paragraph{Results.}For Jacobi iterations, the feedforward difference can accurately trace errors between the current and final samples, which is thus a good metric for convergence and early stopping. In contrast, feedforward differences for the standard feedforward computation are not indicative of convergence. In terms of wall-clock time, Jacobi method only requires \textbf{0.013s} to converge on MNIST, while feedforward computation needs \textbf{0.343s}. This amounts to a speedup factor around \textbf{26}. For CIFAR-10, the time difference is \textbf{0.119s} vs. \textbf{3.026s}, which implies a speedup factor around \textbf{25}. The significant speedup achieved by Jacobi methods for MADE is highly practical. It not only accelerates image generation, but can also directly improve the speed for other models where MADE sampling is a sub-process, such as computing the likelihood of Inverse Autoregressive Flows~\citep{kingma2016improved}, and sampling from Masked Autoregressive Flows~\citep{papamakarios2017masked}.

\subsubsection{PixelCNN++}

PixelCNN++ is a more advanced autoregressive model that typically achieves higher likelihood on image modeling tasks compared to MADE. In addition to the vanilla Jacobi method, we test the proposed hybrid methods, Jacobi-GS and GS-Jacobi.
Feedforward sampling from PixelCNN++ can be accelerated by caching~\citep{ramachandran2017fast}, where the computation performed for one state is memorized to accelerate the computation of later states. As discussed in \secref{sec:which}, parallel Jacobi updates cannot leverage these caches for faster sampling, and therefore one parallel update can be slower than one sequential update of feedforward sampling. Jacobi-GS and GS-Jacobi, in contrast, can take advantage of the caching mechanism since they incorporate sequential updates.

\paragraph{Setup.} We use PixelCNN++ models trained on MNIST and CIFAR-10 datasets. Each experiment is performed 10 times and we show both mean and standard deviation in \figref{fig:pixelcnn_mnist} and \ref{fig:pixelcnn_cifar}. We consider feedforward sampling with and without caches. We implement Jacobi iterations in the same way as MADE, where no cache is used. We modify the caching mechanisms from \cite{ramachandran2017fast} so that they can be applied to Jacobi-GS and GS-Jacobi approaches. For GS-Jacobi, one block contains 15 rows of pixels on MNIST and 2 rows of pixels on CIFAR-10. For Jacobi-GS, one block has one row of pixels on both datasets. All results of wall-clock time are measured on a single NVIDIA Tesla V100 GPU with 32 GB memory. The batch sizes are 16 and 4 for MNIST and CIFAR-10 respectively. %

\begin{table}[t]
    \caption{Speedups for PixelCNN++ sampling on MNIST and CIFAR-10. Algorithms are stopped when the $\ell_\infty$ norm between the current sample and the ground-truth image is smaller than $0.01$ (when the difference in samples is imperceptible to human eyes). }\label{tab:pixelcnn_eps}
    \begin{center}
        \begin{adjustbox}{max width=\linewidth}
        \begin{tabular}{c|cc||cc}
        \Xhline{3\arrayrulewidth} \bigstrut
            \multirow{2}{*}[-3pt]{Method} & \multicolumn{2}{c||}{MNIST} & \multicolumn{2}{c}{CIFAR-10}\\\cline{2-5}
             & Time (s) & Speedup & Time (s) & Speedup \bigstrut \\
             \Xhline{1\arrayrulewidth}\bigstrut
             Feedforward w/o cache & 12.15 & 1.00$\times$ & 30.95 & 1.00$\times$ \\\bigstrut
             Feedforward w/ cache & 8.23 & 1.48$\times$ & 17.76 & 1.74$\times$\\
             \Xhline{1\arrayrulewidth}\bigstrut
             Jacobi & 1.94 & 6.26$\times$ & 26.16  & 1.18$\times$\\\bigstrut
             GS-Jacobi & \textbf{1.86} & \textbf{6.53$\times$} & 14.84 & 2.09$\times$\\\bigstrut
             Jacobi-GS & 5.95 & 2.04$\times$ & \textbf{14.76} & \textbf{2.10}$\times$\\
           \Xhline{3\arrayrulewidth} 
        \end{tabular}
        \end{adjustbox}
    \end{center}
\end{table}

\begin{figure}[th]
  \centering
    \includegraphics[width=\linewidth]{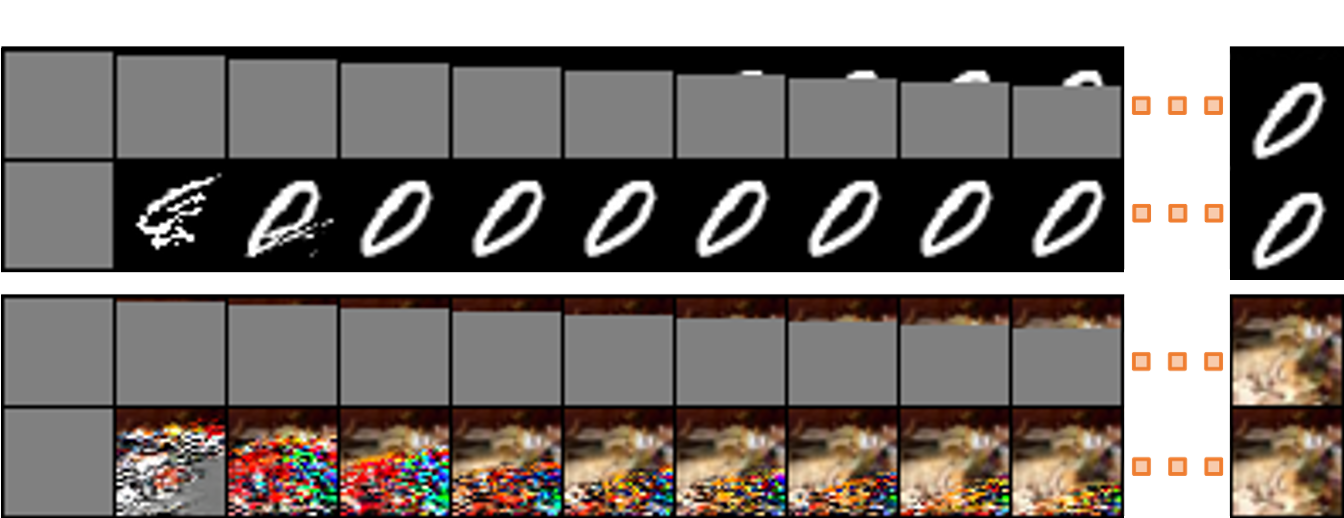}
    \caption{Feedforward (1st \& 3rd rows) vs. Jacobi (2nd \& 4th rows) sampling for PixelCNN++ on MNIST (\textbf{top 2 rows}) and CIFAR-10 (\textbf{bottom 2 rows}). 
    Each column corresponds to the same number of updates.
    We show the first few intermediate samples on the left and the final image samples on the rightmost.}
  \label{fig:pixelcnn_demo_placeholder}
\end{figure}

\paragraph{Results.} 
We report the performance of different samplers in \tabref{tab:pixelcnn_eps}, and include a visual comparison of Jacobi iteration vs. feedforward sampling (\ie, the standard ancestral sampling) in Fig. \ref{fig:pixelcnn_demo_placeholder}. Compared to the standard feedforward computation (ancestral sampling) without caching, Jacobi, Jacobi-GS and GS-Jacobi all run significantly faster. Even against feedforward sampling + caching, our GS-Jacobi and Jacobi-GS methods still perform uniformly better. Specifically, GS-Jacobi yields \textbf{6.53} and \textbf{2.09} times speedup (on MNIST and CIFAR-10) compared to the vanilla feedforward sampling without caching, and yields \textbf{4.42} and \textbf{1.20} times speedup against feedforward sampling + caching. Similarly, Jacobi-GS leads to speedup factors of \textbf{2.04} and \textbf{2.10} compared to the vanilla feedforward sampling, and still have speedup factors of \textbf{1.38} and \textbf{1.20} against feedforward sampling + caching.
Compared to GS-Jacobi, Jacobi-GS may require fewer parallel processing units. For example, Jacobi-GS only requires $28$ parallel computing units on MNIST, since there are $28$ blocks and each block requires only one parallel device to run. In contrast, GS-Jacobi has a block size of $15 \times 28$ and requires the same number of parallel processing units. Our Jacobi method always outperforms the vanilla feedforward sampling without caching, with a speedup factor of \textbf{6.26} and \textbf{1.18} on MNIST and CIFAR-10 respectively. However, as demonstrated by our results on CIFAR-10 (see \tabref{tab:pixelcnn_eps}), Jacobi iterations may become slower than hybrid methods since the latter can exploit caching.

\section{Related Work}
Accelerating feedforward computation in the context of autoregressive sampling has been studied in the literature. In particular, \citet{oord2018parallel} propose probability density distillation to distill information from a slow autoregressive model to a faster sampler. However, it may provide samples from a different distribution compared to the original (slower) autoregressive model. MintNet~\citep{song2019mintnet} proposes a fixed-point iteration method based on Newton-Raphson to speed up the inversion of an autoregressive procedure, but it is limited to a particular model. Similar ideas have also been proposed as a theoretical possibility in \cite{naumov2017parallel} without experimental verifications.

Concurrently, \citet{wiggers2020predictive} propose to accelerate autoregressive sampling with a fixed-point iteration method and demonstrate advantages over feedforward sampling (without caching) on PixelCNN++ models. Our Jacobi approach in \algref{alg:jacobi} is equivalent to theirs, but we additionally provide hybrid methods to improve the vanilla Jacobi approach, which are able to outperform feedforward sampling with caching. Our approaches are also more general, applicable to tasks beyond autoregressive sampling such as RNN training and DenseNet inference.

Common iterative solvers for linear equations include Jacobi, Gauss-Seidel, successive over-relaxation (SOR), and more general Krylov subspace methods. Forward/back substitution, as a process of solving lower/upper triangular linear systems, can also be viewed as instances of feedforward computation. Many approaches are proposed to accelerate and parallelize this procedure. Specifically, level scheduling~\citep{saad2003iterative} performs a topological sorting to find independent groups of variables that can be solved in parallel. Block-Jacobi iteration methods~\citep{anzt2015iterative, anzt2016domain, chow2018using}, similar to the Jacobi-GS method in our paper, are proposed to maximize the parallel efficiency on GPUs.

Jacobi-type iterations are also used in message passing algorithms for probabilistic graphical models~\citep{elidan2012residual,niu2011tuffy} and graph neural networks (GNNs,~\citet{scarselli2008graph}). In particular, Gaussian belief propagation (GaBP) includes the Jacobi method as a special case~\citep{bickson2008gaussian} when solving Gaussian Markov random fields. The core computation of GNNs is a parameterized message passing process where methods similar to block-Jacobi scheduling are popular~\citep{liao2018graph}.

\section{Conclusion}
By interpreting the feedforward computation as solving a triangular system of nonlinear equations, we show that numerical solvers can, in some cases, provide faster evaluation at the expense of additional parallel computing power. In particular, we demonstrated that variants of Jacobi and Gauss-Seidel iterations are effective in accelerating the training of RNNs, the evaluation of DenseNets on ImageNet and the sampling from multiple deep autoregressive models, such as MADE and PixelCNN++, on several image datasets.

This observation opens up many new possible directions. We can build highly-optimized software packages to automatically parallelize some feedforward computation. More sophisticated numerical equation solving techniques, such as Krylov subspace methods and continuation methods, may provide greater acceleration than Jacobi or our hybrid methods. Our idea is particularly useful in time-critical applications, where trading parallel computing power for time is otherwise impossible.

Finally, we reiterate that our method is not beneficial for all feedforward computation. We require the process to tolerate numerical errors, have long skip connections, as well as have weak dependencies among various sequential stages that might be leveraged by numerical solvers (see the discussions in \secref{sec:which}). Moreover, in some cases, it can be non-trivial for practical implementations to reap the benefits of acceleration that are possible in theory due to various overheads in software or hardware. %

\subsection*{Acknowledgements}
This research was supported by Intel Corporation, TRI, NSF (\#1651565, \#1522054, \#1733686), ONR  (N00014-19-1-2145), AFOSR (FA9550-19-1-0024). Yang Song was supported by the Apple PhD Fellowship in AI/ML.

\bibliography{fast}
\bibliographystyle{icml2021}

\newpage
\onecolumn
\appendix
\newpage
\section{Examples of Feedforward Computation}
\label{app:examples}
Feedforward computation is ubiquitous in machine learning. Below we focus on three prominent examples that appear in our experiments.
\subsection{Evaluating Neural Networks}
Suppose we have an input $\bfx$ and a neural network of $L$ layers defined by $f(\bfx) \triangleq a^{L}(\bfb^{L} + \bfW^{L} a^{L-1}(\cdots a^{1}(\bfb^{1} + \bfW^{1} \bfx)))$, where $a^{\ell}(\cdot)$, $\bfb^{\ell}$ and $\bfW^{\ell}$ denote the activation function, bias vector and weight matrix for the $\ell$-th layer respectively. We typically evaluate $f(\bfx)$ via feedforward computation, as can be seen by letting $T = L$, $\bfu = \bfx$, and defining $\bfs_t \triangleq a^{t}(\bfb^{t} + \bfW^{t} a^{t-1}(\bfb^{t-1} + \bfW^{t-1}a^{t-2}(\cdots)))$, $h_1(\bfu) \triangleq a^{1}(\bfb^{1} + \bfW^{1}\bfu)$ and $h_t(\bfu, \bfs_{1:t-1}) \triangleq a^{t}(\bfb^{t} + \bfW^{t} \bfs_{t-1})$ in \eqnref{eqn:feedforward}. By changing $\bfu$, we can evaluate the neural network for different inputs.

\subsection{Backpropagation}
Consider the same neural network as discussed above. Let $\mathbf{r}_\ell \triangleq a^{\ell}(\bfb^{\ell} + \bfW^{\ell} a^{\ell-1}(\bfb^{\ell-1} + \bfW^{\ell-1}a^{\ell-2}(\cdots a^{1}(\bfb^{1} + \bfW^{1} \bfx)))))$, and $\mbf{r}_0 \triangleq \bfx$. Suppose the loss function is $\mcal{L}(\mathbf{r}_L)$. Through the chain rule, we can compute the gradient of $\mcal{L}$ \wrt the $\ell$-th layer by $\nabla_{\mathbf{r}_\ell} \mcal{L} = \mcal{L}'(\mbf{r}_L)$ if $\ell = L$ and $\nabla_{\mathbf{r}_\ell} \mcal{L} = (\bfW^{\ell})\tran \nabla_{\mbf{r}_{\ell + 1}} \mcal{L} \odot (a^{\ell + 1})'(\bfb^{\ell + 1} + \bfW^{\ell + 1} \mbf{r}_\ell)$ if $\ell < L$, where $\odot$ denotes the element-wise product. The backpropagation algorithm for computing $\nabla_{\bfx} \mcal{L}$ can be viewed as feedforward computation, because we can define $\bfu = \{\mbf{r}_0, \mbf{r}_1, \cdots, \mbf{r}_L \}$, and let $T = L+1$, $\bfs_{t} \triangleq \nabla_{\mbf{r}_{L-t + 1}}\mcal{L}$, $h_1(\bfu) \triangleq \mcal{L}'(\mbf{r}_L)$, $h_t(\bfu, \bfs_{1:t-1}) \triangleq (\bfW^{L-t+1})\tran \bfs_{t-1} \odot (a^{L-t + 
2})'(\bfb^{L-t+2} + \bfW^{L-t+2} \mbf{r}_{L-t+1})$ in \eqnref{eqn:feedforward}. Note that the gradients of $\mcal{L}$ \wrt model parameters can be immediately computed after $\bfs_{1:T+1}$ has been obtained.

\subsection{Sampling from Autoregressive Models}\label{sec:autoregressive}
Autoregressive models define a high-dimensional probability distribution $p(\bfx)$ via the chain rule $p(\bfx) = \prod_{i=1}^N p(x_i | x_{1:i-1})$. We can draw samples from this distribution using a sequential process called ancestral sampling. Concretely, we first draw $\tilde{x}_1 \sim p(x_1)$, and then $\tilde{x}_t \sim p(x_t | \tilde{x}_{1:t-1})$ for $t=2,3,\cdots,N$ successively. Let $\bfu = (u_1, u_2, \cdots, u_N)$ denote the states of the pseudo-random number generator that correspond to samples $\tilde{x}_1, \tilde{x}_2, \cdots, \tilde{x}_N$. For example, $u_1, u_2, \cdots, u_N$ may be uniform random noise used in inverse CDF sampling. The ancestral sampling process is an instance of feedforward computation, as in \eqnref{eqn:feedforward} we can set $T = N$, $\bfs_t = \tilde{x}_t$, and let $h_t(\bfu, \bfs_{1:t-1})$ be the pseudo-random number generator that produces $\tilde{x}_t$ from $p(x_t \mid \tilde{x}_{1:t-1})$ given $\bfu$. We can randomly sample the input $\bfu$ to generate different samples from the autoregressive model.

\section{Proofs}
Here we provide the convergence proofs for Jacobi, Jacobi-GS and GS-Jacobi algorithms.
\begin{customprop}{\ref{prop:jacobi}}
\algref{alg:jacobi} converges and yields the same result as standard feedforward computation in at most $T$ parallel iterations for any initialization of $\bfs_{1:T}^0$ if $\epsilon = 0$.
\end{customprop}
\begin{proof}
We prove the conclusion by induction, and without loss of generality we assume the algorithm terminates at the $T$-th iteration. Suppose the true solutions for \eqnref{eqn:triangular} are $\bfs_1^*, \bfs_2^*, \cdots, \bfs_T^*$. For the first parallel iteration, we have $\bfs_1^1 \gets h_1(\bfu) = \bfs_1^*$. Now we hypothesize that for the $k$-th ($k \geq 1$) parallel iteration, $ \forall j \leq k: \bfs_j^k = \bfs_j^*$. Suppose the hypothesis for $k$ is true. Considering the $(k+1)$-th iteration, we have $\bfs_{k+1}^{k+1} \gets h_{k+1}(\bfu, \bfs_{1:k}^{k}) = h_{k+1}(\bfu, \bfs_{1:k}^*) = \bfs_{k+1}^*$. In addition, for $i < k+1$, we have $\bfs_{i}^{k+1} \gets h_{i}(\bfu,\bfs_{1:i-1}^{k}) = h_{i}(\bfu, \bfs_{1:i-1}^*) = \bfs_i^*$. Therefore, we have proved that the hypothesis holds true for $k+1$. Since we have shown that the hypothesis is true for $k=1$, by induction it is true for all $k \geq T$, which implies $\bfs_{1:T}^T = \bfs_{1:T}^*$. In other words, the algorithm gives the true solutions to \eqnref{eqn:triangular} in at most $T$ parallel iterations.
\end{proof}

\begin{customprop}{\ref{prop:others}}
For any initialization, Jacobi-GS (\algref{alg:jacobi_gs}) and GS-Jacobi (\algref{alg:gs_jacobi}) converge in at most $M$ block-wise iterations and yield the same results as obtained by standard feedforward computation if $\epsilon = 0$.
\end{customprop}

\begin{proof}
We first prove the convergence of Jacobi-GS. Suppose the true solutions are $\bfs_1^*, \bfs_2^*, \cdots, \bfs_T^*$, and without loss of generality the algorithm terminates at $k=M$. For the first parallel iteration, we consider block $\mcal{B}_1 = \llbracket a_1, b_1 \rrbracket$. After completing all the GS steps for the first parallel iteration, it is easy to see that $\forall i \in \llbracket a_1, b_1 \rrbracket: \bfs_i^1 = \bfs_i^*$. Now we hypothesize that after the $k$-th ($k \geq 1$) parallel iteration, $\forall t \leq k, \forall i \in \mcal{B}_t: \bfs_i^k = \bfs_i^*$. Consider the $(k+1)$-th iteration. Note that for all $i \leq k+1$, we have $\forall j \in \mcal{B}_i=\llbracket a_i, b_i \rrbracket: \bfs_j^{k+1} \gets h_j(\bfu, \bfs_{1:a_i-1}^{k}, \bfs_{a_i:j-1}^{k+1}) = h_j(\bfu, \bfs_{1:a_i-1}^*, \bfs_{a_i:j-1}^{k+1})$, and GS iterations make sure that $\forall j \in \mcal{B}_i: \bfs_j^{k+1} = \bfs_j^*$. This proves that the hypothesis is true for $k+1$. Since we have shown the correctness of the hypothesis for $k=1$, by induction we know the hypothesis holds true for all $1\leq k \leq M$. This implies that $\bfs_{1:T}^M = \bfs_{1:T}^*$.

Next, we prove the convergence of GS-Jacobi. For the first GS iteration, we know $\forall j \in \mcal{B}_1: \bfs_j^{|\mcal{B}_1|} = \bfs_j^*$ from \propref{prop:jacobi}, and therefore $\bfs_{\mcal{B}_1} = \bfs_{\mcal{B}_1}^{|\mcal{B}_1|} = \bfs_{\mcal{B}_1}^*$. We can simply continue this reasoning to conclude that $\forall 1\leq i \leq M: \bfs_{\mcal{B}_i} = \bfs_{\mcal{B}_i}^{|\mcal{B}_i|} = \bfs_{\mcal{B}_i}^*$. 

\end{proof}

\section{Extra Experimental Details}\label{app:exp_details}
\subsection{RNN}
We train a standard one-layer RNN on resized MNIST images. MNIST is dataset of hand-written digits with 50000 training data and 10000 test data. The original resolution of these images is $28\times 28$, and we resize it to $10\times 10$. The RNN has 128 hidden units, and uses the \verb|SoftPlus| activation function. All weights are initialized with Gaussian noise of mean zero and standard deviation $0.1$. The bias parameters are initialized to zero. We train the RNN with stochastic gradient descent, where the learning rate is 0.0001, batch size is 1, and momentum is 0. All experiments are implemented with PyTorch and run on an Nvidia Titan Xp GPU. All GPU timing is done after properly calling \verb|torch.cuda.synchronize()|.

\subsection{DenseNets}
We use the DenseNet-201 model provided by the PyTorch~\citep{paszke2019pytorch} model zoo, which has been pre-trained on the ImageNet~\citep{ILSVRC15} dataset with a top-5 error of 6.43\%. We use an Nvidia Titan Xp GPU in our experiments. All GPU timing is done after properly calling \verb|torch.cuda.synchronize()|.

\subsection{MADE}

Our MADE network has two layers, each with 512 neurons. For training MADE on both MNIST and CIFAR-10, we use a batch size of 128, a learning rate of 0.001 for the Adam optimizer, and a step-wise learning rate decay of 0.999995. The models were trained for 1000 epochs. During sampling, we produce 100 images in parallel from our MADE model. We use a logistic distribution to model each conditional probability. For MNIST images, the resolution is $28\times28\times 1$ and thus $T=784$. For CIFAR-10 images, the resolution is $32\times 32 \times 3$ and therefore $T=3072$. We use a single Nvidia Titan Xp GPU for our experiments and measure wall-clock time after properly calling \verb|torch.cuda.synchronize()|.

\subsection{PixelCNN++}

For CIFAR-10, we use the same architecture and checkpoint provided by the original PixelCNN++ paper~\citep{salimans2017pixelcnn++}. For MNIST, the architecture is the same as that for CIFAR-10, except that we shrink the number of filters to $1/4$. We train the models on MNIST using a batch size of 32, a learning rate of 0.0002 for the Adam optimizer, and a step-wise learning rate decay of 0.999995. The model was trained for 590 epochs. All models are implemented in JAX~\citep{jax2018github} and FLAX~\citep{flax2020github}.

For MNIST, the image resolution is $28\times28\times 1$ and therefore $T=784$. For CIFAR-10 images, the resolution is $32\times32\times 3$, but different from MADE, PixelCNN++ views all channels at one location as one state, which means $T = 32\times 32 = 1024$. In our experiments, we run everything on an Nvidia Tesla V100 GPU (32 GB). All GPU timing is done by calling \verb|.block_until_ready()| properly.

\section{Extra Experimental Results}\label{app:exp_results}

\begin{figure}
    \centering
    \subfigure[Before training\label{fig:rnn_before}]{\includegraphics[width=0.4\textwidth]{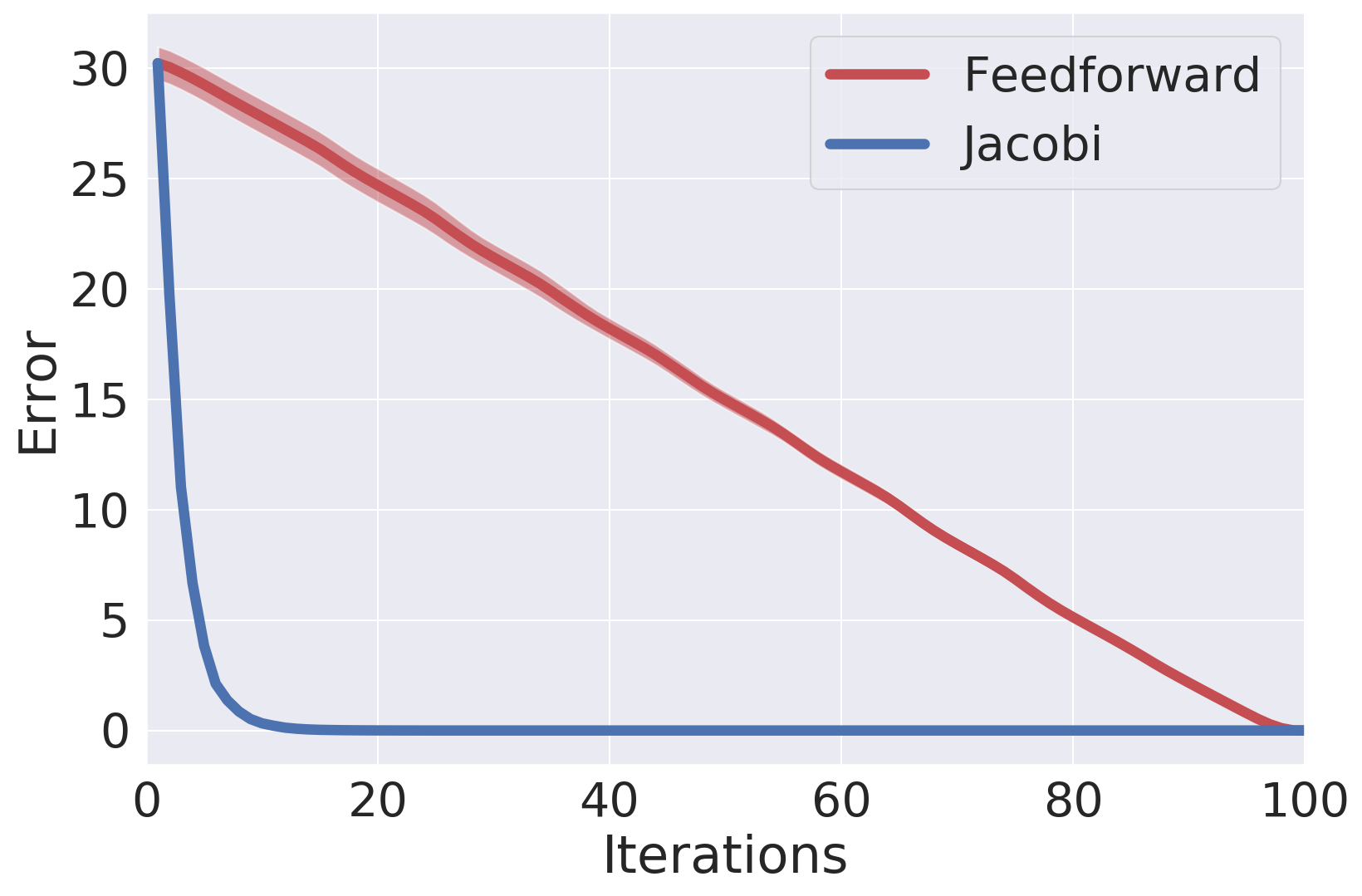}}
    \subfigure[After training\label{fig:rnn_end}]{\includegraphics[width=0.4\textwidth]{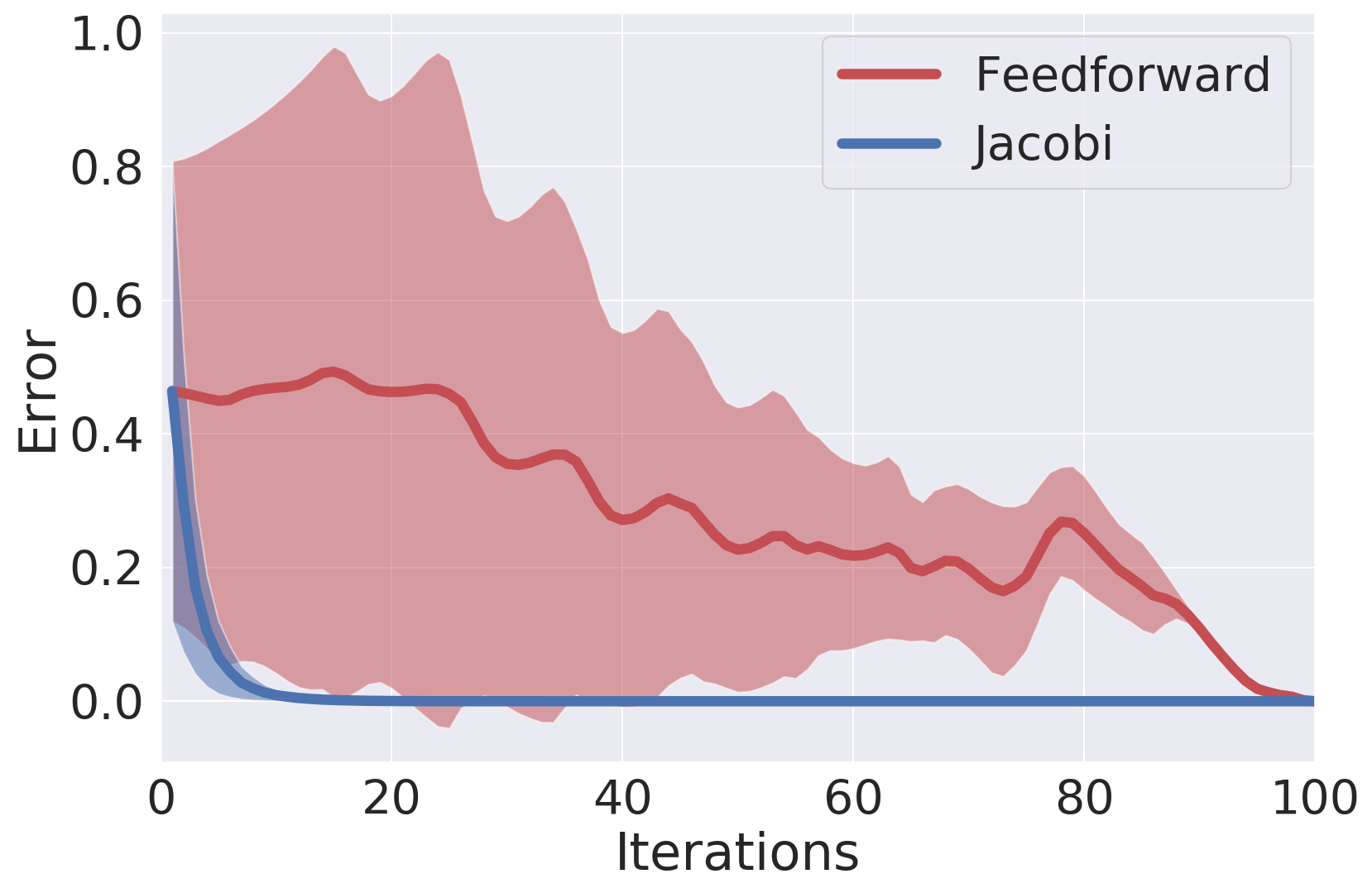}}
    \caption{Convergence of gradient errors when using Jacobi iterations to accelerate the backpropagation of RNNs. Gradient errors are measured with $\ell_2$ norm, and averaged over 10 runs. The shaded area denotes $\nicefrac{1}{10}$ of standard deviations.}
    \label{fig:rnn_convergence}
\end{figure}

\begin{figure}
    \centering
    \includegraphics[width=0.6\linewidth]{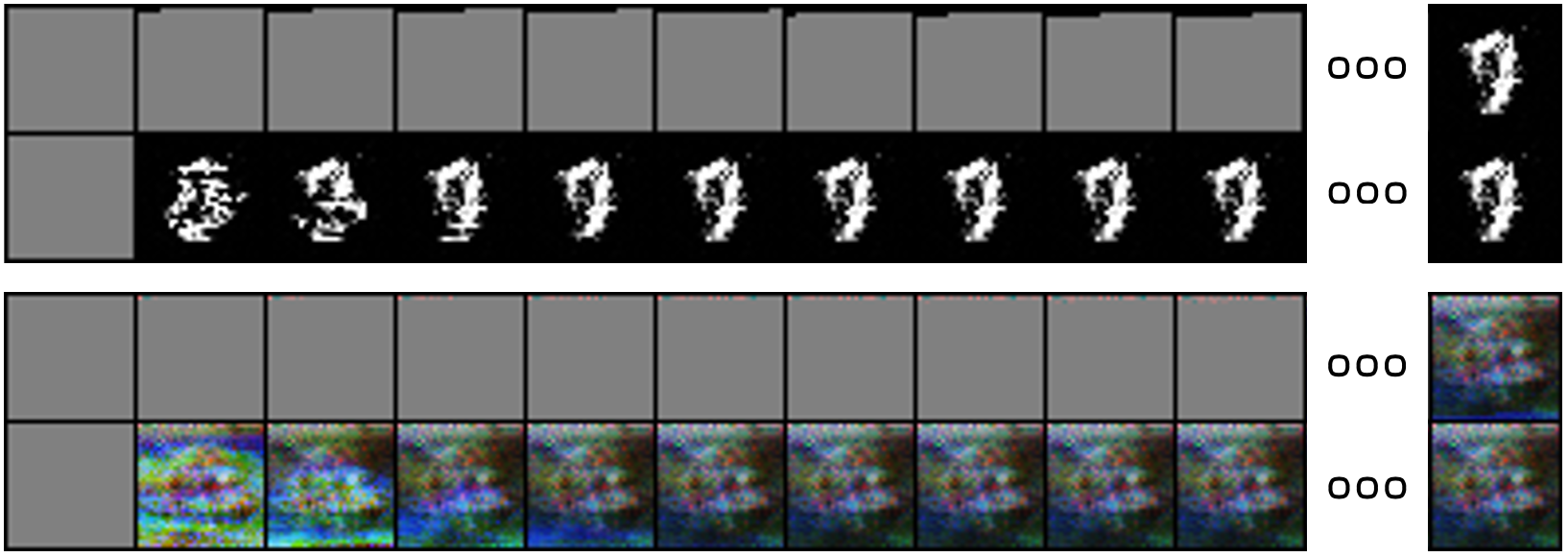}
    \caption{Demonstration of the Jacobi sampling process for MADE on MNIST (\textbf{top two rows}) and CIFAR-10 (\textbf{bottom two rows}). The odd rows correspond to standard feedforward sampling, and the even rows are from the Jacobi sampling process. We show the intermediate samples every five (parallel) iterations on the left side of the ellipses, and the final image samples on the right.}
    \label{fig:made_demon}
\end{figure}

\begin{figure}
    \centering
    \subfigure[RNN training]{\includegraphics[width=0.34\linewidth, valign=c]{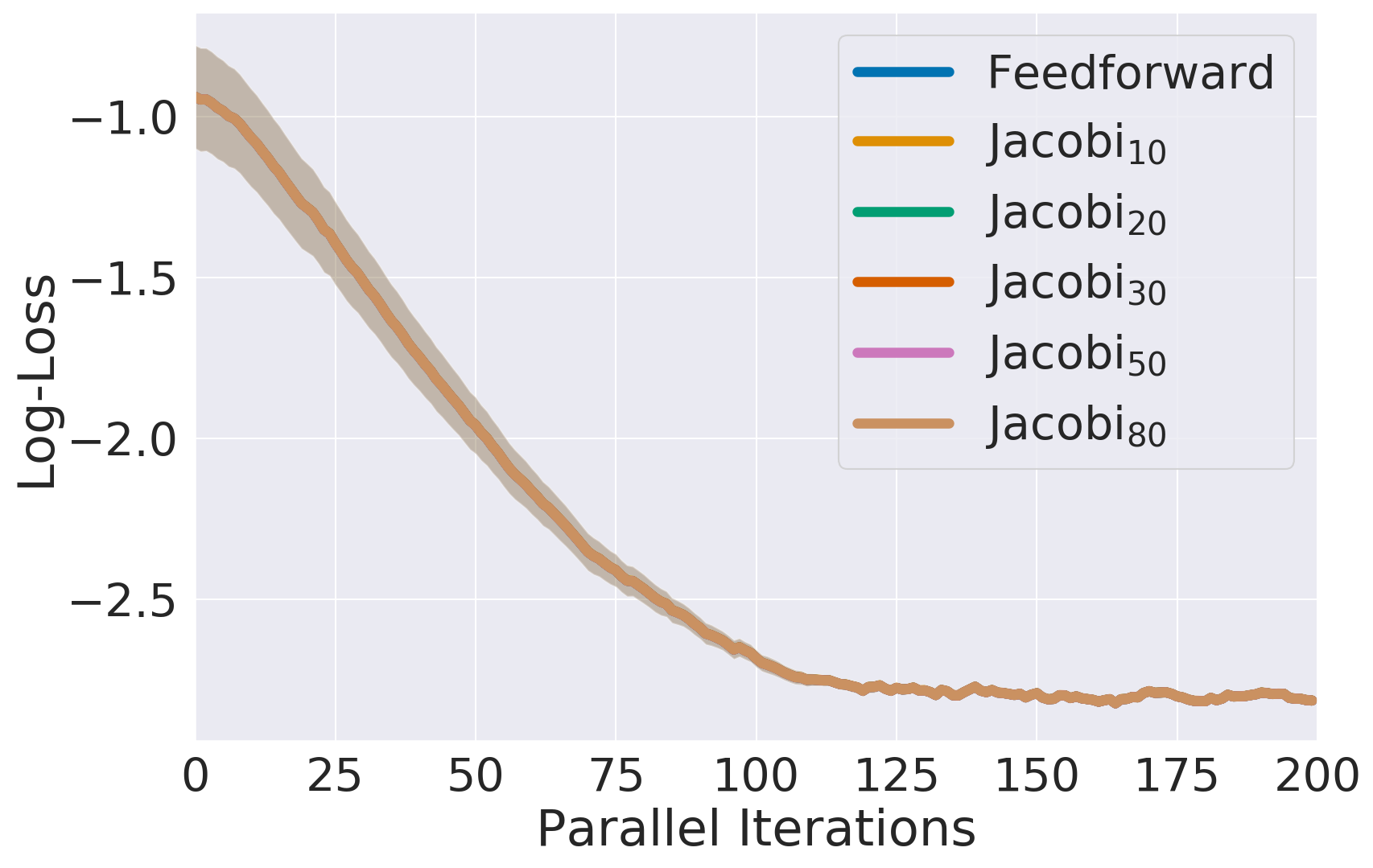}}%
    \subfigure[MADE sampling on MNIST]{\includegraphics[width=0.325\linewidth, valign=c]{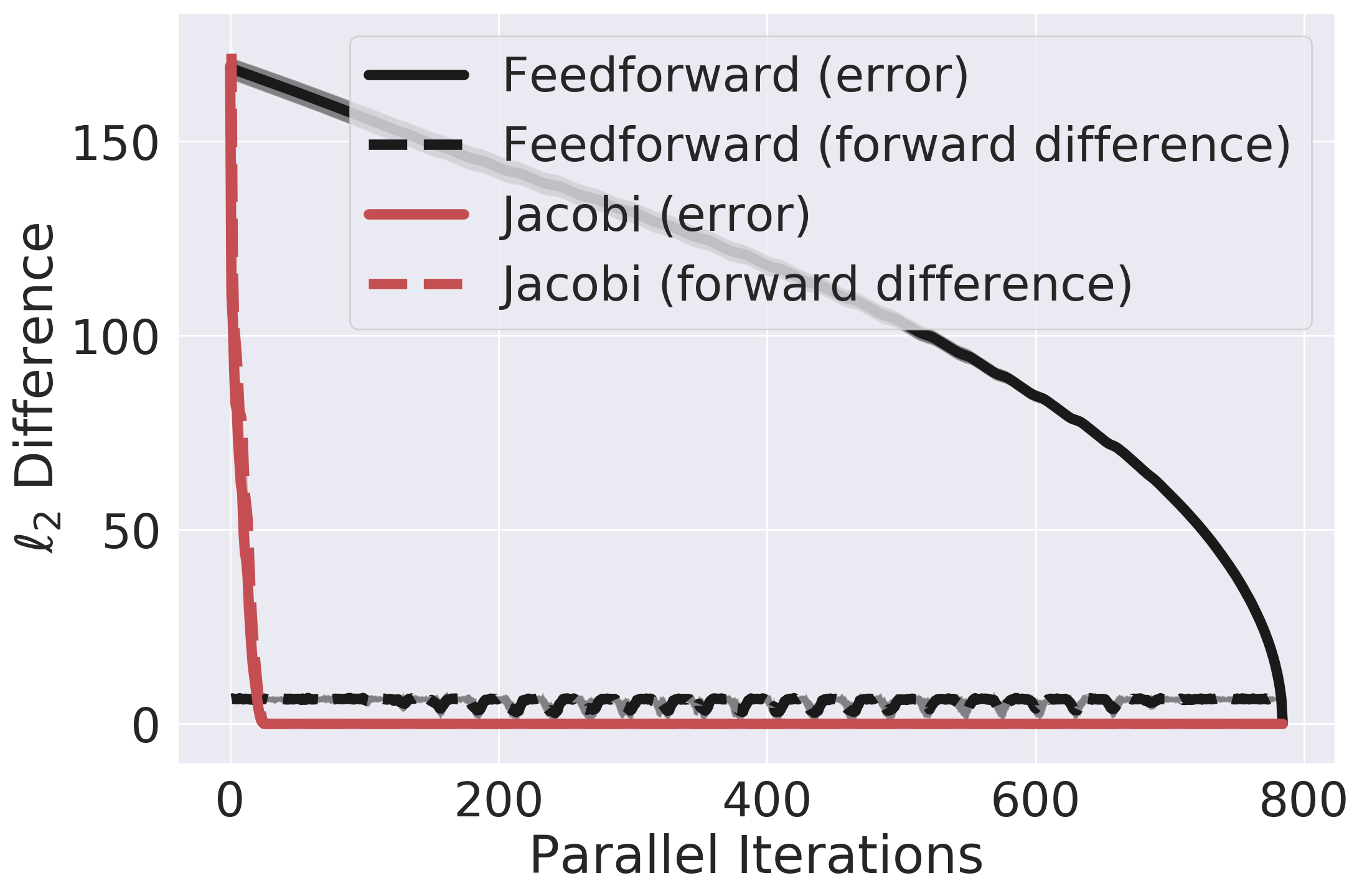}}
    \subfigure[PixelCNN++ sampling on MNIST]{\includegraphics[width=0.325\linewidth, valign=c]{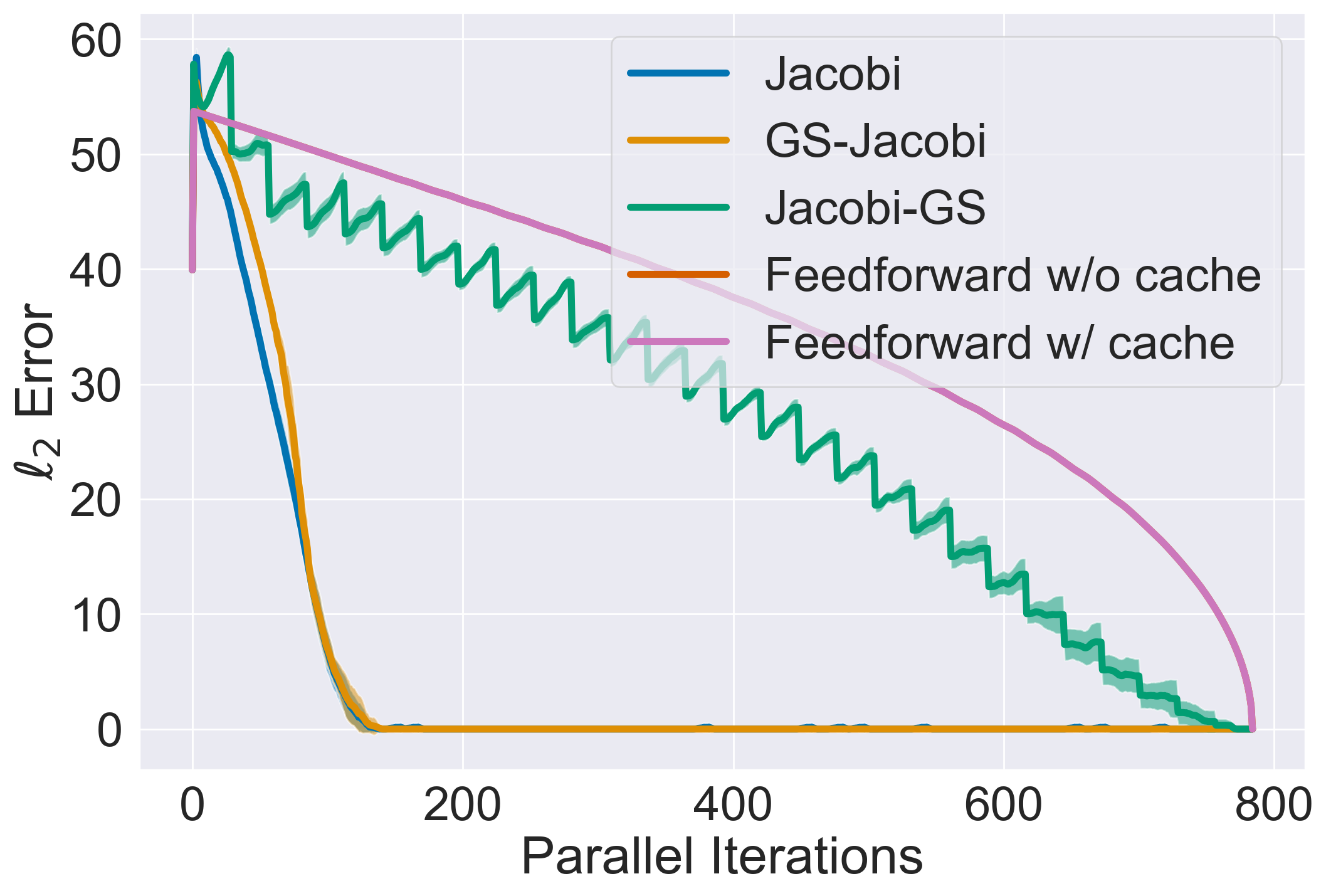}}\\
    \subfigure[DenseNet evalution]{\includegraphics[width=0.325\linewidth, valign=c]{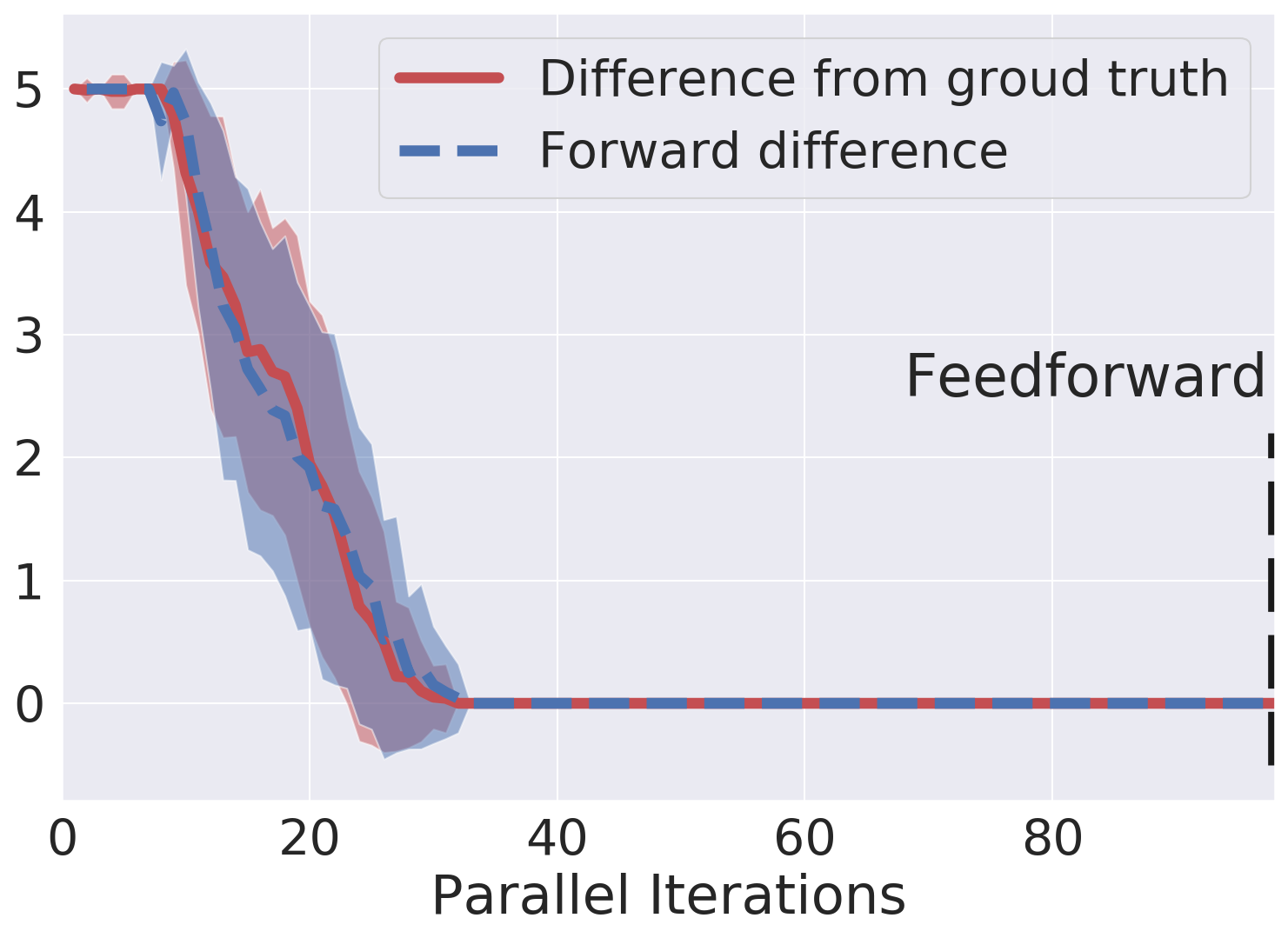}}
    \subfigure[MADE sampling on CIFAR-10]{\includegraphics[width=0.338\linewidth, valign=c]{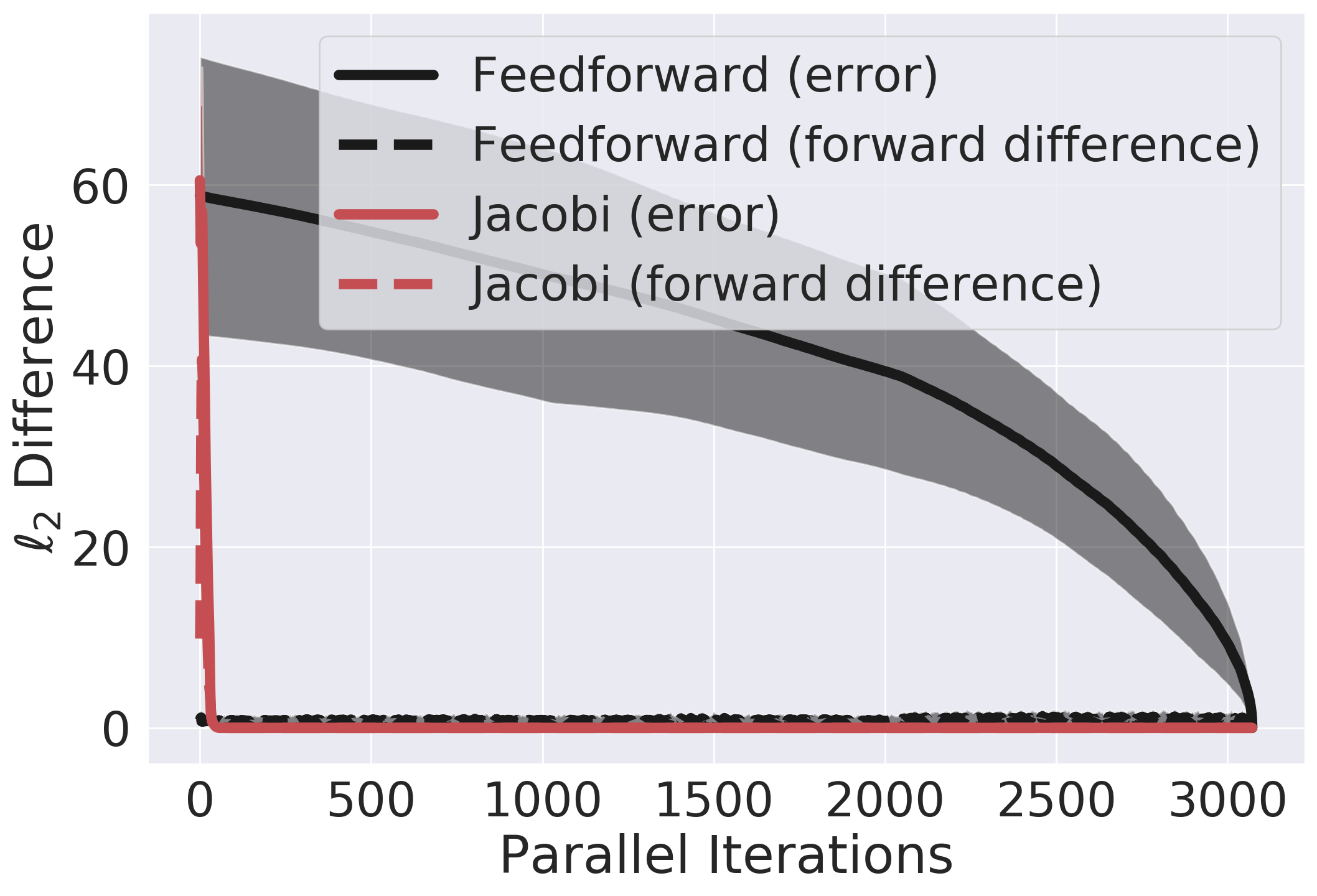}}%
    \subfigure[PixelCNN++ sampling on CIFAR-10]{\includegraphics[width=0.33\linewidth, valign=c]{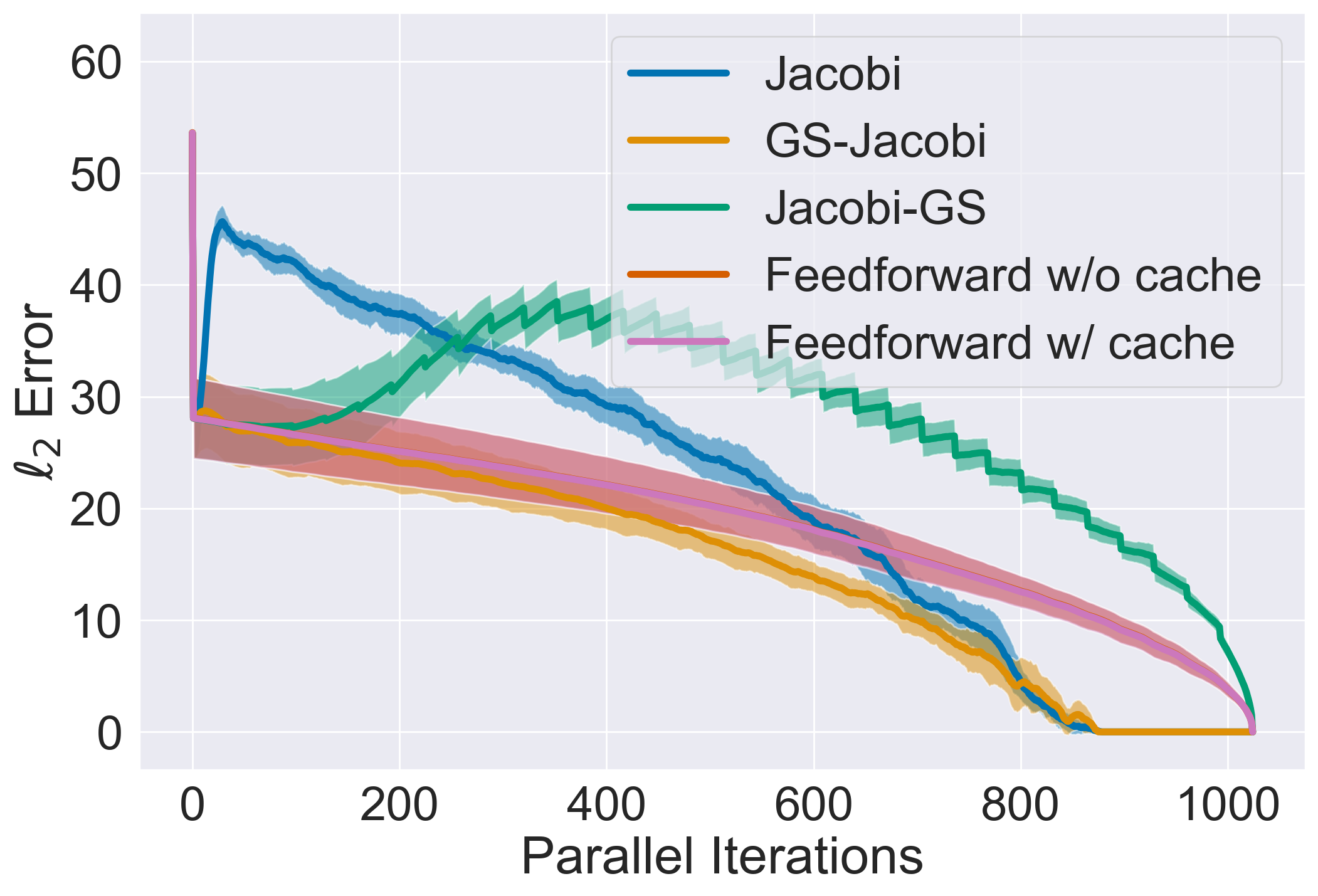}}

    \caption{(a) The performance of Jacobi iterations on accelerating RNN training. Here we use ``Jacobi$_n$'' to denote the Jacobi method truncated at the $n$-th iteration, and ``feedforward'' for standard backpropagation. All values are averaged over 10 runs and shaded areas represent $\nicefrac{1}{10}$ of standard deviations. All curves coincide with each other. (d) The performance of Jacobi-GS on evaluating DenseNets. The y-axis represents the number of incorrect labels in top-5 predictions. The shaded areas represent standard deviations across 100 random input images. (b)(e) The performance of feedforward sampling vs. Jacobi iterations for MADE. The shaded areas represent standard deviations computed over 100 runs. (c)(f) Comparing different sampling algorithms for PixelCNN++. Results are averaged over 10 runs and shaded areas show standard deviations.}\label{fig:iters}
\end{figure}

We provide additional results on how fast the Jacobi algorithm converges for RNN backpropagation. Note that the performance of Jacobi methods will change gradually as the RNN model parameters evolve during training. We report the convergence results before training starts in \figref{fig:rnn_before}, and after training finishes in \figref{fig:rnn_end}, where Jacobi methods have a clear advantage in both cases. We also provide a demonstration on the standard feedforward sampling procedure vs. our Jacobi sampling method for MADE in \figref{fig:made_demon}. In \figref{fig:iters}, we show how various methods convergence with respect to the number of (parallel) iterations in lieu of the wall-clock time (\cf, \figref{fig:results}).

\end{document}